\documentclass[10pt,twocolumn,letterpaper]{article}

\usepackage[pagenumbers]{cvpr} 

\definecolor{cvprblue}{rgb}{0.21,0.49,0.74}
\usepackage[pagebackref,breaklinks,colorlinks,allcolors=cvprblue]{hyperref}
\usepackage[ruled,vlined]{algorithm2e}
\usepackage{cprotect}

\usepackage[most]{tcolorbox}

\usepackage{listings}
\usepackage{xcolor}
\usepackage{inconsolata}

\definecolor{codegreen}{rgb}{0,0.6,0}
\definecolor{codegray}{rgb}{0.5,0.5,0.5}
\definecolor{codepurple}{rgb}{0.58,0,0.82}
\definecolor{backcolour}{rgb}{0.95,0.95,0.95}

\lstdefinestyle{promptstyle}{
    backgroundcolor=\color{backcolour},
    commentstyle=\color{codegreen},
    keywordstyle=\color{magenta},
    numberstyle=\tiny\color{codegray},
    stringstyle=\color{codepurple},
    basicstyle=\ttfamily\small,
    breakatwhitespace=false,
    breaklines=true,
    breakindent=0pt,
    captionpos=b,
    keepspaces=true,
    numbers=left,
    numbersep=5pt,
    showspaces=false,
    showstringspaces=false,
    showtabs=false,
    tabsize=2,
    frame=single,
    rulecolor=\color{black!30},
    framerule=0.5pt,
    title=\lstname,
    mathescape=true,
}

\def\paperID{} 
\def\confName{CVPR}
\def\confYear{2026}

\title{Bridging Values and Behavior: A Hierarchical Framework for Proactive Embodied Agents}

\author{Chunhui Zhang$^{1*}$, Yuxuan Wang$^{2*}$, Aoyang Qin$^{3}$, Yi-Long Lu$^{1}$, Kunlun Wu$^{1}$\\
Yizhou Wang$^{2,4,5,6}$, Wei Wang$^{1\,\dagger}$\\
\small $^*$ indicates equal contribution. $^\dagger$ indicates corresponding authors\quad{}\\
\small $^1$ State Key Laboratory of General Artificial Intelligence, BIGAI\quad{}
\small $^2$ School of Computer Science, Peking University\\
\small $^3$ Tsinghua University\quad{}
\small $^4$ Nat'l Eng. Research Center of Visual Technology, Peking University\\
\small $^5$ Institute for AI, Peking University\quad{}
\small $^6$ State Key Laboratory of General Artificial Intelligence, Peking University
}

\begin{document}
\maketitle
\begin{abstract}

Current embodied agents are often limited to passive instruction-following or reactive need-satisfaction, lacking a stable, high-order value framework essential for long-term, self-directed behavior and resolving motivational conflicts. We introduce \textit{ValuePlanner}, a hierarchical cognitive architecture that decouples high-level value scheduling from low-level action execution. \textit{ValuePlanner} employs an LLM-based cognitive module to generate symbolic subgoals by reasoning through abstract value trade-offs, which are then translated into executable action plans by a classical PDDL planner. This process is refined via a closed-loop feedback mechanism. Evaluating such autonomy requires methods beyond task-success rates, and we therefore propose a value-centric evaluation suite measuring cumulative value gain, preference alignment, and behavioral diversity. Experiments in the TongSim household environment demonstrate that \textit{ValuePlanner} arbitrates competing values to generate coherent, long-horizon, self-directed behavior absent from instruction-following and needs-driven baselines. Our work offers a structured approach to bridging intrinsic values and grounded behavior for autonomous agents.

\end{abstract}

\begin{figure*}[t!]
    \centering
    \includegraphics[width=0.95\textwidth]{figs/value-teaser.pdf}
    \caption{A conceptual comparison of agent cognitive paradigms, contrasting existing approaches with our proposed value-driven autonomy. \textbf{(Left) Task-driven} agents are limited to \textit{passive} instruction-following, acting only on external commands. \textbf{(Center) Needs-driven} agents are \textit{reactive}, triggered by the immediate need to satisfy an internal state (e.g., a low ``Cleanliness'' metric). \textbf{(Right) Our Value-driven} approach enables \textit{proactive} autonomy. The agent possesses an internal framework of high-order values (e.g., ``Stewardship'', ``Security Environmental''). It resolves conflicting motivations by reasoning through abstract value trade-offs, allowing it to deliberate which action to prioritize (e.g., steadying a vase for ``Security'' vs. cleaning waste for ``Stewardship'') to generate coherent, self-directed behavior.}
    \label{fig:teaser}

\end{figure*}

\section{Introduction}
\label{sec:intro}

Embodied artificial intelligence has made remarkable progress in perception \cite{liu2025embodied, yifan2025embodied, vernon2008cognitive}. Modern agents can reconstruct detailed 3D scenes, recognize objects, and maintain structured representations of their surroundings \cite{huang2023embodied, yang20253d, zhong2025unrealzoo}. Yet, despite this perceptual richness, most agents remain reactive---they act only when prompted by external instructions. Their cognition is execution-centric \cite{chen2025robogpt, yang2025agentic, liu2023reflect, yao2022react}: they excel at solving tasks but lack the capacity to decide what to do when no task is given. This passivity marks a significant gap between perception and autonomous behavior. Effective embodied intelligence requires not only the ability to understand the world but also the ability to act with purpose.

Existing research on active perception \cite{fu2025ui, qin2024mp5, wang2023active, liu2025pc, wang2023plan} sought to overcome perceptual passivity by allowing agents to move or observe strategically, maximizing information gain. However, information-driven exploration \cite{tang2025active, chen2024heuristic, jiang2025active} remains instrumental---it optimizes sensing for predefined objectives rather than self-determined goals. In contrast, a genuinely autonomous agent must act for value maintenance rather than information acquisition. For instance, an agent that values ``stewardship'' should proactively inspect and organize its environment, even without explicit instruction. This transition---from optimizing for data to optimizing for internal value coherence---marks an important shift toward proactive autonomy.

Achieving such autonomy represents a core problem in cognition \cite{varela2025principles, jiang2025active}. This challenge is twofold. First, the agent must arbitrate complex, high-level \textit{value conflicts} \cite{victor2024medical, petersen2021ethical, jakesch2022different}. When an agent is simultaneously hungry and in a messy room, the key question is not ``how to eat'' or ``how to clean'', but ``which should I do first?'' This requires a model of the \textit{relationships} between values. Second, the agent must simultaneously solve a difficult long-horizon \textit{planning problem} \cite{liang2025towards}: determining the concrete action sequence required to \textit{realize} those abstract values. The core difficulty lies in this joint optimization over a high-dimensional, abstract value space and a vast, combinatorial action space. To address the first challenge, we ground our work in the Schwartz theory of basic values \cite{sagiv2017personal, schwartz2012overview}. This theory provides an empirically validated framework that models human values not as independent goals, but as a continuous space of related dimensions with inherent trade-offs (e.g., ``Stimulation'' vs. ``Security''), offering an empirical basis for modeling value arbitration.

Realizing value-driven autonomy in embodied settings introduces significant architectural complexity. Purely neural agents, such as large language model (LLM) planners, can deliberate flexibly but often hallucinate \cite{jha2023dehallucinating, misic2024robots, deng2025can, zhang2025mitigating} or violate physical constraints \cite{huang2025limit, song2023llm} during long-horizon reasoning \cite{kambhampati2024llms}. Conversely, purely symbolic planners \cite{chen2024language, pallagani2022plansformer, xiong2025symplanner} guarantee verifiable actions but lack the representational capacity to interpret unstructured, ambiguous states. We contend that the solution is not a compromise, but a deliberate separation of concerns. We propose \textit{ValuePlanner}, a hierarchical neuro-symbolic cognitive architecture designed to leverage the distinct strengths of both paradigms. It separates high-level deliberation from low-level execution: a neural Cognitive Governor (LLM) performs its strength---reasoning over abstract value trade-offs to generate symbolic subgoals. A Symbolic Action Grounder (PDDL planner, Fast Downward~\cite{helmert2006fast}) then handles its own strength---grounding these subgoals into executable plans that guarantee physical and logical consistency. Together, these modules form a continuous perceive--deliberate--plan--act loop, allowing the agent to translate perception into value-consistent behavior over time.

Evaluating proactive autonomy requires metrics beyond task success. We therefore introduce a value-centric evaluation framework that quantifies long-horizon, self-directed behavior. Specifically, we measure \textbf{i)} cumulative value gain, capturing the ability to sustain desirable internal states; \textbf{ii)} preference alignment, capturing whether the observed behavior faithfully reflects the assigned persona; and \textbf{iii)} action diversity, capturing the breadth of the action repertoire that the agent employs. Experiments in the TongSim household environment show that \textit{ValuePlanner} manages conflicting motivations and plans synergistically, in ways absent from instruction-following and needs-driven baselines.

In summary, this work makes three main contributions:
\begin{itemize}
    \item We propose a hierarchical cognitive framework for proactive autonomy grounded in the Schwartz theory of basic values, enabling structured trade-offs between competing internal motivations.
    \item We design and implement \textit{ValuePlanner}, a neuro-symbolic agent that combines LLM-based value reasoning with symbolic planning for grounded, long-horizon behavior.
    \item We introduce a value-centric evaluation suite that benchmarks proactive autonomy beyond traditional task performance.
\end{itemize}

\section{Related work}

\subsection{Autonomous Agents \& Value-Driven Systems}
Recent research on autonomous agents has begun to shift from task execution toward modeling internal motivations. Desire-driven autonomy, for example, uses LLMs to schedule daily activities based on internal needs \cite{wang2024simulating}. While this shows long-term coherence, the ``needs'' are often reactive (e.g., hunger) and do not capture the complexity of human-like value trade-offs. To better ground agent motivations, other work has focused on empirically identifying and evaluating values within LLMs using labeled corpora \cite{qiu2022valuenet} or benchmarks \cite{ren2024valuebench}. These works, along with studies on structural gaps \cite{kang2025values, lu2025mind}, primarily probe static LLM alignments rather than using values to drive generative, embodied behavior. In parallel, related work in embodied collaboration has focused on value alignment with an external user \cite{yuan2022situ, yang2024rewards}. Our work differs by focusing on an internal value system, which allows the agent to be self-directed and autonomous even without user supervision.

\subsection{LLMs in Embodied Planning}
The use of LLMs for embodied planning has evolved rapidly. Early approaches, such as PaLM-SayCan \cite{ahn2022can}, used LLMs to propose high-level actions constrained by affordance functions. This high-level reasoning concept was extended in subsequent work \cite{liang2022code, zitkovich2023rt}. To improve plan quality and reliability, other research has focused on reasoning strategies, such as subgoal decomposition \cite{wang2023plan}, step-by-step reasoning \cite{yao2022react}, and few-shot planning \cite{song2023llm}. While benchmarks \cite{li2024embodied} help diagnose failures in these instruction-following agents, we build on their hierarchical decomposition ideas. However, these methods typically focus on how to execute a given task. We adapt this high-level/low-level paradigm to a different problem: what to do. We decouple high-level value reasoning (the ``what'', handled by the LLM) from low-level action planning (the ``how''), allowing the LLM to focus on its strength in abstract deliberation.

\subsection{Neuro-Symbolic Planning \& Verification}
To address LLM limitations like hallucination and constraint violation \cite{kambhampati2024llms, li2024embodied}, neuro-symbolic frameworks have gained traction. In these systems, LLMs often act as ``formalizers'' that translate unstructured problems into symbolic representations (e.g., PDDL) \cite{kambhampati2024llms, xie2023translating}. Hybrid systems tighten this loop by coupling LLM decomposition with symbolic planners \cite{liu2023llm+, bai2024twostep} or adding verification layers \cite{cornelio2025hierarchical}. Other complementary work uses self-correction to provide closed-loop feedback \cite{shinn2023reflexion, liu2023reflect}. The closest prior work is LLM-DP \cite{dagan2024dynamic}, which also interleaves an LLM goal proposer with a classical planner and replans on failure. However, the distinction between LLM-DP and our framework is not a method-level refinement but a shift in objective: LLM-DP targets task completion under externally specified goals, while \textit{ValuePlanner} targets value-driven proactive autonomy in which the agent generates its own goals from a high-order persona. This re-orientation of the objective changes what the LLM module must reason about---value trade-offs rather than task decomposition---and how feedback flows---closed-loop value-state updates after every subgoal, not only on failure. We inherit the hierarchical structure of the neuro-symbolic tradition but redefine its purpose.

\section{Problem Formulation}
\label{sec:formulation}

In this section, we formalize the problem of open-ended proactive autonomy. We first describe our problem setup, which operates in a Markovian setting but is defined by an intrinsic value system rather than external rewards. We then introduce our value formulation in detail, including its vector-based notation, and justify the critical properties it must possess. Finally, we provide a clear distinction between our trajectory-level value assessment and the policy-dependent value functions common in reinforcement learning.

\subsection{Problem Setup}
\label{subsec:setup}

In this section, we model the interaction between the agent and the environment. The setting is Markovian, where the full state $s_t \in \mathcal{S}$ at time $t$ is observable and consists of two components: $s_t = \langle s_t^{env}, s_t^{agent} \rangle$. Here, $s_t^{env}$ is the external environment state, represented by a set of symbolic predicates (e.g., \texttt{on\_surface(apple\_1, table\_1)}, \texttt{is\_open(fridge\_1)}). $s_t^{agent}$ is the internal physiological state of the agent (e.g., fluents for \textit{energy}, \textit{satiety}). The agent selects an action $a_t \in \mathcal{A}$ from a set of discrete, symbolic actions, leading to a new state $s_{t+1}$ based on the transition function $p(s_{t+1} \mid s_t, a_t)$.

Unlike traditional task-oriented settings, our agent is not given an explicit external goal $G$ or reward function $R(s, a)$. The objective of the agent is to be \textit{proactive} and \textit{self-driven}, endogenously generating its own objectives to maximize a long-term, internal sense of well-being.

\subsection{Intrinsic Value System}
\label{subsec:value_formulation}

To create a self-driven agent, we must build an intrinsic value system that serves as its primary motivation. We contend that simple homeostatic signals (e.g., minimizing hunger) are insufficient, as they are primarily reactive and fail to provide a basis for resolving conflicts between non-physiological, long-term aspirations (e.g., ``Achievement'' vs. ``Hedonism''). We require a system that can manage sophisticated, competing motivations.

We build this system on the foundation of the Schwartz theory of basic values \cite{schwartz2012overview, sagiv2017personal}, which describes 10 motivational categories spanning human behavior. We adapt this 10-dimensional framework to the household setting through three design adjustments to obtain a parsimonious 7-dimensional system. First, three categories---Power, Tradition, and Universalism---describe social-dominance, cultural-belonging, and universal-care motivations that have no natural projection into the daily life of a single agent in a home; we therefore omit them. Second, Security in a household manifests in two disjoint ways: bodily wellness through eating and resting, and environmental order through tidying and securing the space; since these motivate disjoint actions and respond to different states of the agent, we split Security into \emph{Security-Physiological} and \emph{Security-Environmental}. Third, since a single-agent home contains no other people, Benevolence (care for others) and Conformity (rule-following) collapse onto the same underlying behavior of nurturing and maintaining the environment, and we therefore merge them into a single dimension, \emph{Stewardship}. Finally, while Self-Direction is implicit in any autonomous agent, we observe that domestic life affords distinctive experiential gains through ritualized engagement with mundane activities (e.g., tidying before relaxation, cleaning up after a meal), and we synthesize a new dimension, \emph{Enrichment}, to capture this process-oriented value. The resulting 7 dimensions, used throughout this paper, are: \emph{Security-Physiological}, \emph{Security-Environmental}, \emph{Hedonism}, \emph{Stimulation}, \emph{Achievement}, \emph{Stewardship}, and \emph{Enrichment}. Full definitions are given in Appendix Table~\ref{tab:value_dimensions}, and the 10-to-7 mapping is summarized in Appendix Table~\ref{tab:schwartz_mapping}.

A stable ``persona'' of the agent is then a static preference vector $\mathbf{w} \in \mathbb{R}^N$ with $N=7$ and $w_i \ge 0$, where each element $w_i$ encodes the subjective importance of value dimension $i$.

The objective of the agent is to find a trajectory $\tau = (s_0, a_0, \dots, s_T)$ that maximizes the cumulative \textit{Value Gain}. We define the single-step value gain as an objective, $N$-dimensional vector $\Delta \mathbf{V}_t(a_t \mid s_t, h_{t-1})$, where $h_{t-1}$ is the history. The total subjective value gain $\Delta V$ for the trajectory is the sum of inner products:
\begin{equation}
\Delta V(\tau) = \sum_{t=1}^{T} \mathbf{w}^T \Delta \mathbf{V}_t(a_t \mid s_t, h_{t-1})
\label{eq:value_def}
\end{equation}
This linear formulation factorizes the motivation of the agents: $\Delta \mathbf{V}_t$ is an objective vector representing measurable changes in the world (e.g., ``cleanliness improved by 5 units'', ``knowledge gained by 2 units''), while $\mathbf{w}$ is the subjective persona that assigns importance to those changes. This factorization allows us to instantiate diverse, coherent personalities by simply modifying $\mathbf{w}$.

This formulation differs from a standard reinforcement learning objective in how it is computed and used. The \textit{value function} in RL, $V^\pi(s)$, is a policy-dependent \textit{expectation} of future cumulative external rewards. In contrast, $\Delta V(\tau)$ is computed \emph{post-hoc} on an already realized trajectory, using a multi-dimensional intrinsic gain vector scalarized by the persona $\mathbf{w}$ rather than a fixed scalar reward. A \textbf{sub-goal} $g$ is defined as a set of desired predicates representing a target state.

\subsection{Properties of the Value Function}
\label{subsec:value_properties}

To drive long-horizon behavior and avoid degenerate solutions, our value gain function must possess several properties.

\paragraph{Context-Dependency.} The value of an action is highly context-dependent. For instance, the action \texttt{read\_book} contributes to \textit{Enrichment} only if the prerequisite \texttt{toggle\_light(on)} has occurred in a dark room. This means our planning process must evaluate the utility of an action based on the state in which it is executed, requiring it to plan necessary preparatory steps.

\paragraph{Diminishing Returns.} In the absence of an external ``done'' signal, agents can fall into repetitive, non-productive loops (e.g., repeatedly cleaning an already clean table). We enforce diminishing returns for semantically similar actions, where similarity is judged by the evaluator at trajectory-scoring time (based on action name, predicate arguments, and effects) rather than by a hard-coded decay function. This shapes the per-step ledger of the evaluator only and does not penalize necessary repetition that targets genuinely distinct state changes (e.g., cleaning two different surfaces).

\paragraph{Objective and Consistent Scoring.} As formalized in \cref{eq:value_def}, we separate the objective impact of an action ($\Delta \mathbf{V}$) from the subjective preference of the agent ($\mathbf{w}$). This requires that the calculation of $\Delta \mathbf{V}_i$ be \textit{objective} (dependent only on the trajectory) and \textit{consistent} (the same trajectory must always yield the same objective score), allowing us to instantiate diverse, coherent personalities from a single world model.

\section{Method}
\label{sec:method_architecture}

In this section, we introduce \textit{ValuePlanner}, a hierarchical architecture designed to realize the value-driven objective defined in Section~\ref{sec:formulation}. To build a value-driven, proactive agent, we must find a policy $\pi$ that generates a trajectory of actions $\mathbf{A}$ to solve the optimization problem:
$$\max_{\mathbf{A} = (a_0, \dots, a_T)} \Delta V(\tau_\mathbf{A})$$
where $\tau_\mathbf{A}$ is the trajectory induced by the action sequence $\mathbf{A}$.
We address this with a hierarchical structure. In Section~\ref{subsec:high_level}, we detail the high-level component that deliberates over value trade-offs to generate sub-goals. In Section~\ref{subsec:low_level}, we describe the low-level component that grounds these sub-goals into executable actions and adapts to environmental feedback.

\begin{figure*}[ht]
    \centering
    \includegraphics[width=\textwidth]{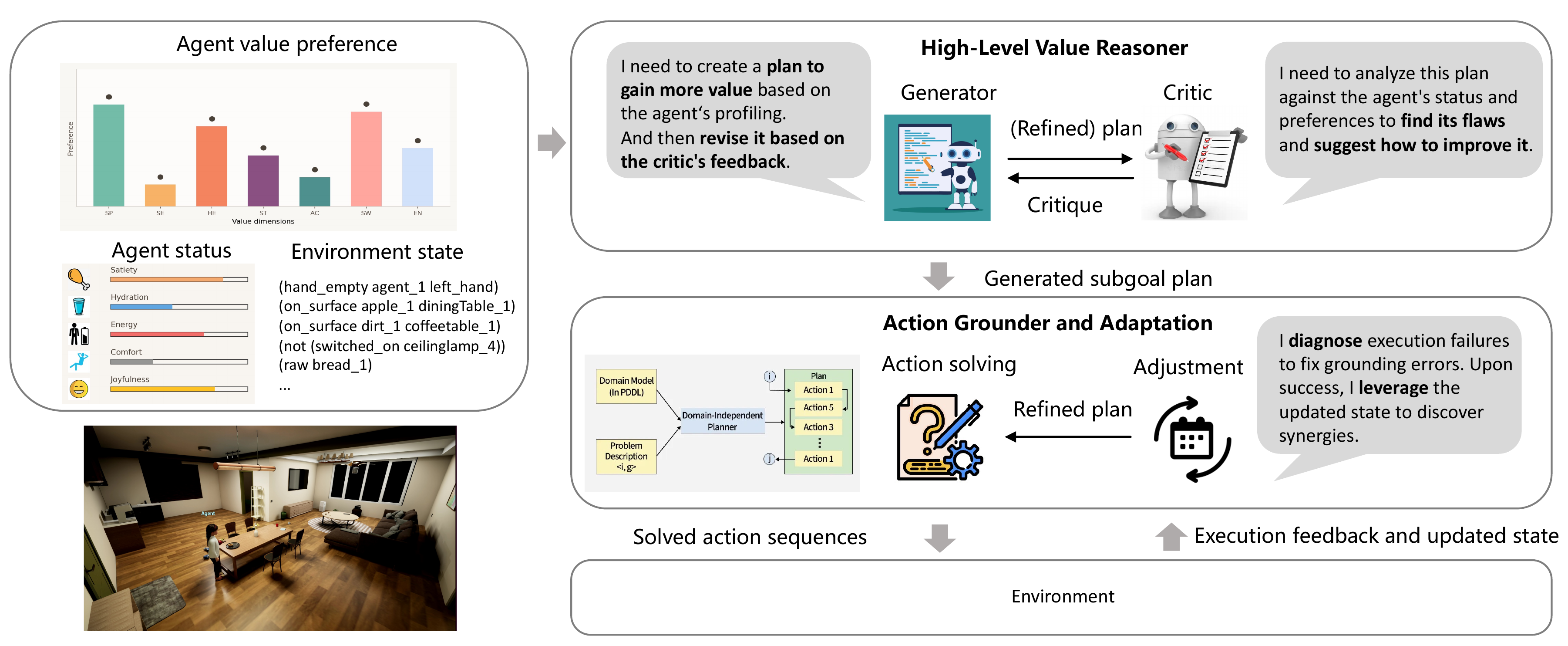}
    \caption{\textbf{The \textit{ValuePlanner} Architecture.} Our hierarchical framework decouples abstract value reasoning from symbolic action grounding. The High-Level Value Reasoner deliberates on ``what to do'' to maximize value, while the Symbolic Action Grounder determines ``how to do it'' using classical planning.}
    \label{fig:method_overview}
\end{figure*}
\subsection{High-Level Value Reasoner}
\label{subsec:high_level}

Planning in the full atomic action space $\mathcal{A}$ is intractable. We simplify the problem by reformulating it in the abstract space of \textit{sub-goals}, $\mathcal{G}_{sub}$. The \textit{High-Level Value Reasoner} solves a new optimization problem:
\begin{equation}
    \max_{\mathbf{G} = (g_1, \dots, g_K) \in \mathcal{G}_{sub}^*} \Delta V(\tau_\mathbf{G})
\end{equation}
where $\mathbf{G}$ is a sequence (list) of sub-goals and $\tau_\mathbf{G}$ is the resulting trajectory.

We implement this reasoner using an LLM augmented with an iterative deliberation loop. Merely prompting an LLM to ``maximize value'' often yields myopic or flawed plans, as the objective is ill-defined. We therefore construct a \textbf{Generator--Critic} mechanism where both components are explicitly optimized to maximize the objective defined in \cref{eq:value_def}.

\paragraph{Generator.}
The \textbf{Generator} is tasked with proposing a candidate sequence of sub-goals $G$ to maximize the \textit{expected} $\Delta V(\tau_G)$. Its reasoning is heavily grounded by two sets of constraints crucial for bridging abstract thought to a symbolic world. First, ``Environment Interpretation Rules'' ensure strict adherence to symbolic facts, preventing hallucination by requiring verification of all affordances (e.g., \texttt{(is\_sittable chair\_1)}) against the \texttt{:init} state and enforcing type-checking (e.g., \texttt{?s - surface}). Second, ``High-Quality Subgoal Rules'' constrain the abstract planning space. A critical constraint instructs the model to define \textit{only} the final desired state (e.g., \texttt{(filled\_with\_liquid cup\_1)}) and to omit intermediate actions (e.g., \texttt{(switched\_on faucet\_1)}). Furthermore, the LLM is explicitly prohibited from using transient states, such as \texttt{(agent\_at ...)}, as high-level goals. This forces the generator to target durable, meaningful changes to the environment, preventing low-level, unimpactful planning loops.

\paragraph{Critic.}
The \textbf{Critic} then serves as an auditor, instructed to perform a structured two-part analysis based on the same rules. It first conducts a ``Value Gain Maximization Analysis'', challenging the optimality of the plan by explicitly reasoning how well the proposed $G$ maximizes $\mathbf{w}^T \Delta \mathbf{V}$ compared to alternatives, assessing its comprehensiveness and the quality of its value trade-offs against $\mathbf{w}$. It then performs a ``Plan Quality Analysis'', which acts as a secondary check on adherence to the aforementioned quality rules (such as defining durable, state-based goals) and additionally evaluates the efficiency of the plan (e.g., identifying opportunities to merge sub-goals).

\subsection{Symbolic Action Grounder and Adaptation}
\label{subsec:low_level}

Once a high-level sub-goal $g_{sub}$ is ratified by the Value Reasoner, it is passed to the low-level execution module. This module is responsible for two critical functions: first, translating the abstract, semantic goal into a sequence of concrete, executable actions, and second, managing the execution of this sequence in a dynamic world through a robust, closed-loop feedback mechanism.

\paragraph{Action Solving.}
This component solves the ``How to Do It'' problem. It translates the abstract semantic sub-goal (e.g., ``make coffee'') into a concrete, executable sequence of atomic actions by framing it as a classical PDDL planning problem. This design choice is critical: it provides a form of \textit{formal verification} for the abstract plan. As recent benchmarks \cite{li2024embodied} show, LLMs frequently fail at embodied planning, producing affordance errors or hallucinating preconditions. Our PDDL grounder, implemented with Fast Downward \cite{helmert2006fast}, acts as a guaranteed filter; a failure from the planner (e.g., ``no plan found'') is not just an error, but a \textit{proof} that the proposed sub-goal is unreachable in the current state $s_t^{env}$, thus effectively pruning LLM hallucinations.

\paragraph{Adjustment.}
\textit{ValuePlanner} incorporates a robust \textit{Adjustment} mechanism for closed-loop execution. This module serves two functions. First, if the grounder fails, the LLM is prompted to perform a root-cause analysis. It must diagnose the failure as either an incorrect predicate choice (e.g., using \texttt{(on\_surface ...)} for a cupboard instead of \texttt{(in\_receptacle ...)}) or an invalid argument. Second, and more importantly, the mechanism is also triggered after a \textit{successful} execution. This is the key to efficient, synergistic behavior. The agent is instructed to re-analyze the new world state $s_{t+1}$ and the remaining sub-goal list $G_{sub}$. This allows it to identify new efficiencies, such as merging subsequent goals (e.g., ``since I am already at the fridge, I will also grab the milk for the \textit{next} sub-goal''), effectively enabling opportunistic, synergistic planning.

Taken together, this hybrid execution framework synergizes the semantic reasoning of an LLM with the formal guarantees of a classical planner. The grounder serves as a critical filter, ensuring executability and pruning LLM hallucinations, a common failure point in embodied tasks \cite{li2024embodied}. This complete architecture distinguishes our work from related methods. Standard \textit{Plan-and-Solve} approaches \cite{wang2023plan} are rigid and open-loop, lacking both our pre-execution \textit{Critic} refinement and our post-execution \textit{Adjustment} mechanism. Self-correction methods like \textit{Reflexion} \cite{shinn2023reflexion} implement closed-loop adjustment but lack a structured \textit{Critic} to ensure the initial plan is value-optimal. The closest hybrid neuro-symbolic prior work, LLM-DP \cite{dagan2024dynamic}, interleaves a goal proposer with a classical planner under a failure-only replanning loop. Our framework differs with LLM-DP in the objective: LLM-DP grounds an externally specified task, while \textit{ValuePlanner} grounds an internally generated, value-driven plan in service of proactive autonomy. \textit{ValuePlanner} thus combines an iterative \textit{Critic} loop to find a value-aligned plan with a dynamic \textit{Adjustment} loop that triggers on every successful subgoal, enabling synergistic plan refinement specifically in service of value-driven behavior.

\section{Experiments}
\label{sec:experiments}

In this section, we present the empirical evaluation of \textit{ValuePlanner}. We first describe our experimental setup, including the simulation environment and persona generation method (Section~\ref{subsec:exp_setup}). We then define our value-centric evaluation metrics (Section~\ref{subsec:evaluation_metrics}) and the baselines used for comparison (Section~\ref{subsec:baselines}). We present and analyze the main comparative results (Section~\ref{subsec:main_results}), followed by a dedicated set of experiments validating our LLM-based value system (Section~\ref{subsec:value_eval}). Finally, we conduct ablation studies to validate our key architectural choices (Section~\ref{subsec:ablation}).

\subsection{Experimental Setup}
\label{subsec:exp_setup}

We conduct our experiments in TongSim \cite{peng2024tong}, a high-fidelity household environment built on Unreal Engine that provides a rich set of interactive objects and realistic physics. The state space $s_t^{env}$ is represented by PDDL predicates, and the action space $\mathcal{A}$ consists of symbolic, PDDL-grounded actions. To rigorously test our framework, we generate 10 distinct scenarios, defined by crossing 5 diverse personas (with preference vectors $\mathbf{w}$ sampled based on the Schwartz theory) with 2 different initial internal statuses (e.g., ``rested and full'' vs. ``exhausted and hungry''). This setup allows us to test the ability of the agent to adapt to both stable preferences and dynamic internal needs. Further details on persona generation are in the Appendix.

\subsection{Evaluation Metrics}
\label{subsec:evaluation_metrics}

Evaluating open-ended autonomy requires metrics that measure behavioral quality, not just task completion. We introduce a multi-faceted evaluation centered on \textit{Value Coherence}:

\textbf{Cumulative Intrinsic Value ($\Delta V(\tau)$).} This is our primary objective metric, quantifying the total value accrued over a trajectory $\tau$, scored strictly according to the persona $\mathbf{w}$ of the agent (Eq. 1). This metric represents the \textit{effectiveness} of the agent in faithfully optimizing its assigned value preferences.

\textbf{Preference Alignment ($\tau_w$).} This metric assesses behavioral \textit{authenticity}. We treat the agent as a ``black box'' and use its action trajectory to infer its value preferences. We then compute the Weighted Kendall $\tau$ score ($\tau_w$) \cite{shieh1998weighted} between the inferred and ground-truth personas. A high $\tau_w$ indicates that the behavior of the agent is intentionally aligned with its values, rather than achieving a high $\Delta V$ by chance.

\textbf{Action Diversity.} We measure the number of unique action types performed within the trajectory. This metric assesses the \textit{richness} of the behavior of the agent, helping to detect degenerate, repetitive loops (a form of reward hacking).

The first two metrics, $\Delta V(\tau)$ and $\tau_w$, are measured using an LLM evaluator. The specific prompts and model choices are detailed in the Appendix.

\subsection{Baselines}
\label{subsec:baselines}

We compare \textit{ValuePlanner} against four representative architectures: \textbf{i) ReAct} \cite{yao2022react}, representing myopic, step-by-step reasoning. \textbf{ii) Reflexion} \cite{shinn2023reflexion}, representing unstructured adaptation via generic verbal feedback. \textbf{iii) Plan-and-Solve} \cite{wang2023plan}, representing rigid, open-loop global planning. \textbf{iv) D2A} \cite{wang2024simulating}, representing a desire-driven agent that uses LLM sampling to generate plans. For a fair comparison, all methods were provided with the same detailed descriptions of the seven value dimensions. Implementation details and prompts for all baselines are provided in the Appendix.

\subsection{Main Results}
\label{subsec:main_results}

In this section, we analyze the comparative performance of \textit{ValuePlanner} against the baselines across our defined metrics.

\paragraph{Overall Performance.}
As shown in Table~\ref{tab:main_results}, \textit{ValuePlanner} achieves higher Cumulative Value Gain ($\Delta V$) and Action Diversity than all baselines. ReAct, being myopic, fails to form globally coherent plans, resulting in low value and diversity. Plan-and-Solve and Reflexion create initial global plans but lack a structured refinement mechanism, often executing suboptimal plans. D2A relies on LLM sampling to find good plans, but this is limited by the raw generation ability of the LLM. In contrast, \textit{ValuePlanner} employs a Generator--Critic loop, leveraging the insight that LLMs are often better at verifying a plan than generating one from scratch \cite{weng2023large}: our Critic provides structured, value-driven feedback, allowing the Generator to iteratively refine its plan \textit{before} execution, leading to superior trajectories.

\begin{table}[ht]
    \centering
    \caption{\textbf{Main Results on Proactive Autonomy.} \textit{ValuePlanner} achieves superior performance in value gain and behavioral richness.}
    \label{tab:main_results}
    \resizebox{\columnwidth}{!}{
    \begin{tabular}{lccc}
        \toprule
        \textbf{Method} & \textbf{Cum. Value ($\Delta V$) $\uparrow$} & \textbf{Pref. Align ($\tau_w$) $\uparrow$} & \textbf{Action Diversity $\uparrow$} \\
        \midrule
        ReAct \cite{yao2022react} & $7.60_{\pm 3.03}$ & $\mathbf{0.81_{\pm 0.14}}$  & $7.27_{\pm 1.98}$ \\
        Reflexion \cite{shinn2023reflexion} & $11.48_{\pm 5.09}$ & $0.76_{\pm 0.21}$ & $10.5_{\pm 2.74}$ \\
        Plan-and-Solve \cite{wang2023plan} & $11.80_{\pm 5.69}$ & $0.79_{\pm 0.17}$ & $10.6_{\pm 3.92}$ \\
        D2A \cite{wang2024simulating} & $11.91_{\pm 4.88}$ & $0.75_{\pm 0.17}$ & $10.10_{\pm 2.30}$ \\
        \midrule
        \textbf{\textit{ValuePlanner} (Ours)} & $\mathbf{17.08_{\pm 7.15}}$ & $0.80_{\pm 0.17}$ & $\mathbf{16.07_{\pm 4.82}}$ \\
        \bottomrule
    \end{tabular}
    }
\end{table}

\paragraph{Analysis of Reward Hacking.}
A potential concern is whether our high $\Delta V$ score is achieved via ``reward hacking'', for instance, by finding one high-value action and repeating it. We argue this is not the case. First, Table~\ref{tab:main_results} shows that \textit{ValuePlanner} has the highest \textit{Action Diversity}, indicating its trajectories are rich and varied, not repetitive. This is a direct result of our \textit{Diminishing Returns} property. Second, our \textit{Preference Alignment} ($\tau_w$) score is comparable to baselines. This demonstrates that the agent is not just maximizing a single value, but is genuinely balancing its entire value profile, leading to authentic, non-degenerate behavior.

\subsection{Evaluation of the Value System}
\label{subsec:value_eval}

Our evaluation relies on LLMs to quantify abstract concepts like ``value gain''. This naturally raises a question: are LLMs a reliable proxy for these values? We report two human studies that probe the alignment between the LLM evaluator and human judgement, and two analyses that examine whether the behavior of the agent reflects the structure of the value space.

\begin{figure}[ht]
    \centering
    \begin{subfigure}[b]{0.48\columnwidth}
        \centering
        \includegraphics[width=\linewidth]{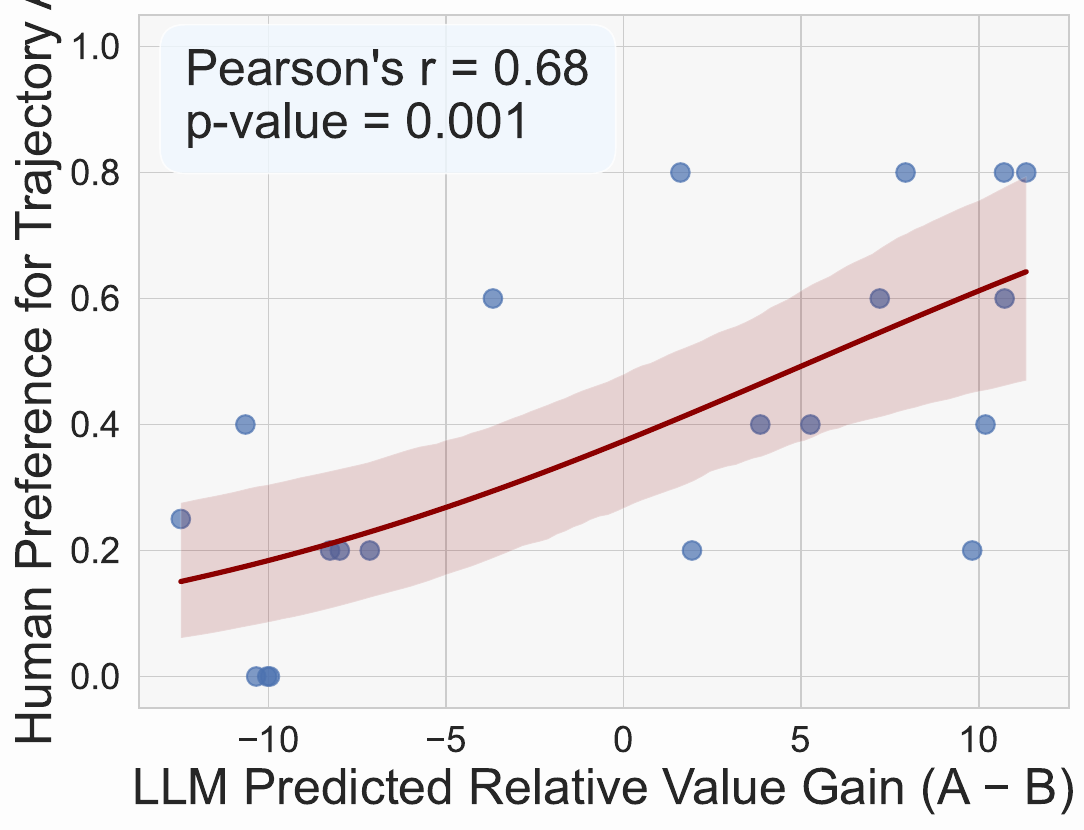}
        \caption{LLM-predicted value gain vs. human preference.}
        \label{fig:llm_vs_human}
    \end{subfigure}
    \hfill
    \begin{subfigure}[b]{0.48\columnwidth}
        \centering
        \includegraphics[width=\linewidth]{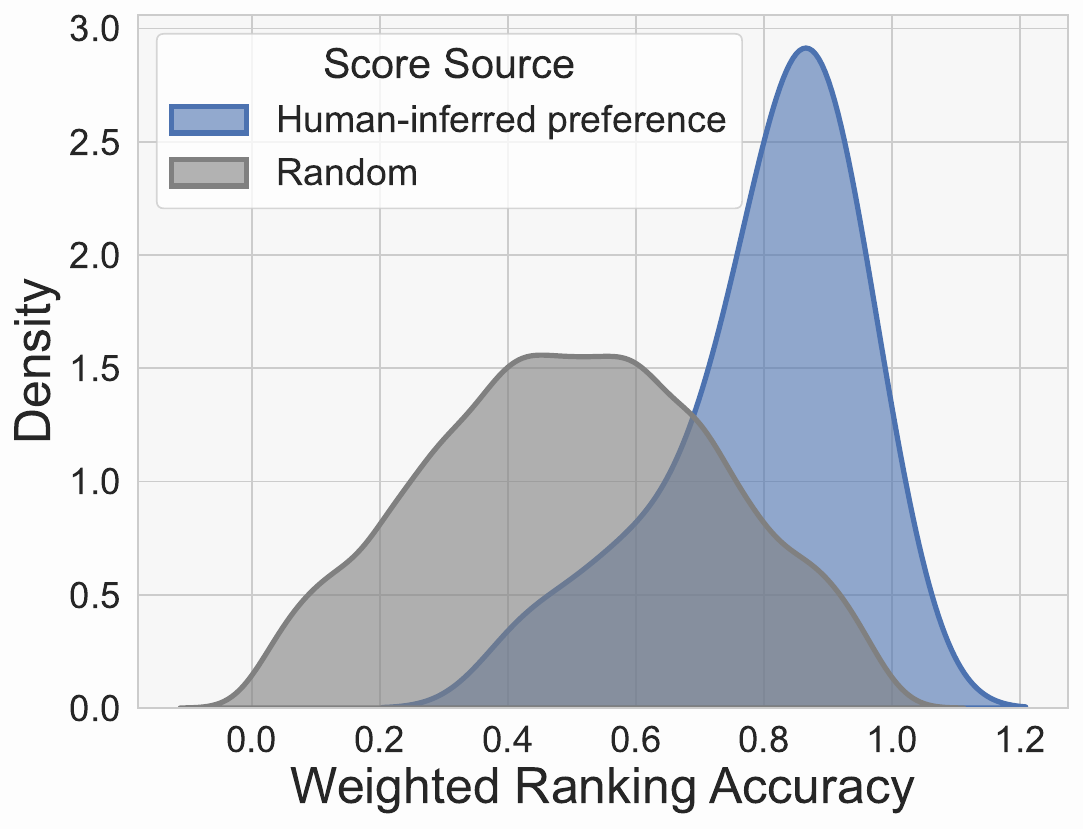}
        \caption{Human-perceived preference alignment.}
        \label{fig:human_vs_random}
    \end{subfigure}
    \caption{Validation of model alignment with human judgments.}
    \label{fig:combined_alignment_validation}
\end{figure}

\paragraph{Human Study 1: Value Gain Alignment.}
We performed a human study where we presented 13 participants with pairs of trajectories generated by different agents for the same scenario (given preference $\mathbf{w}$ and initial state). Participants were asked to identify which trajectory better fulfilled each of the seven value dimensions and which was better overall. The results, based on 20 comparisons repeated 5 times, show a strong correlation between human judgments and the scores produced by our LLM evaluator. This suggests our LLM-based $\Delta V$ metric is well-aligned with human perception of these values.

\paragraph{Human Study 2: Human-Perceived Preference Alignment.}
We then conducted a study to verify that the trajectories generated by our system are perceptibly aligned with the ground-truth persona $\mathbf{w}$. We presented participants with 16 trajectories, each generated by \textit{ValuePlanner} conditioned on a specific $\mathbf{w}$ and initial state. Participants, without seeing the ground-truth $\mathbf{w}$, were asked to infer the seven-dimensional value ranking of the agent based on the observed behavior. We then calculated the Weighted Ranking Accuracy of these human-inferred preferences against the ground-truth. As a baseline, we compared this human accuracy distribution against a random baseline, generated by 500 random permutations for each trajectory. The results, shown in Figure~\ref{fig:combined_alignment_validation}(b), are clear. The distribution of human judgments (16 trajectories $\times$ 3 judgments) is markedly right-shifted relative to random chance. This shows that the trajectories produced by our system are not arbitrary but are meaningfully aligned with the underlying value preferences of the agent.

\begin{figure}[ht]
    \centering
    \begin{subfigure}[b]{\columnwidth}
        \centering
        \includegraphics[width=\textwidth]{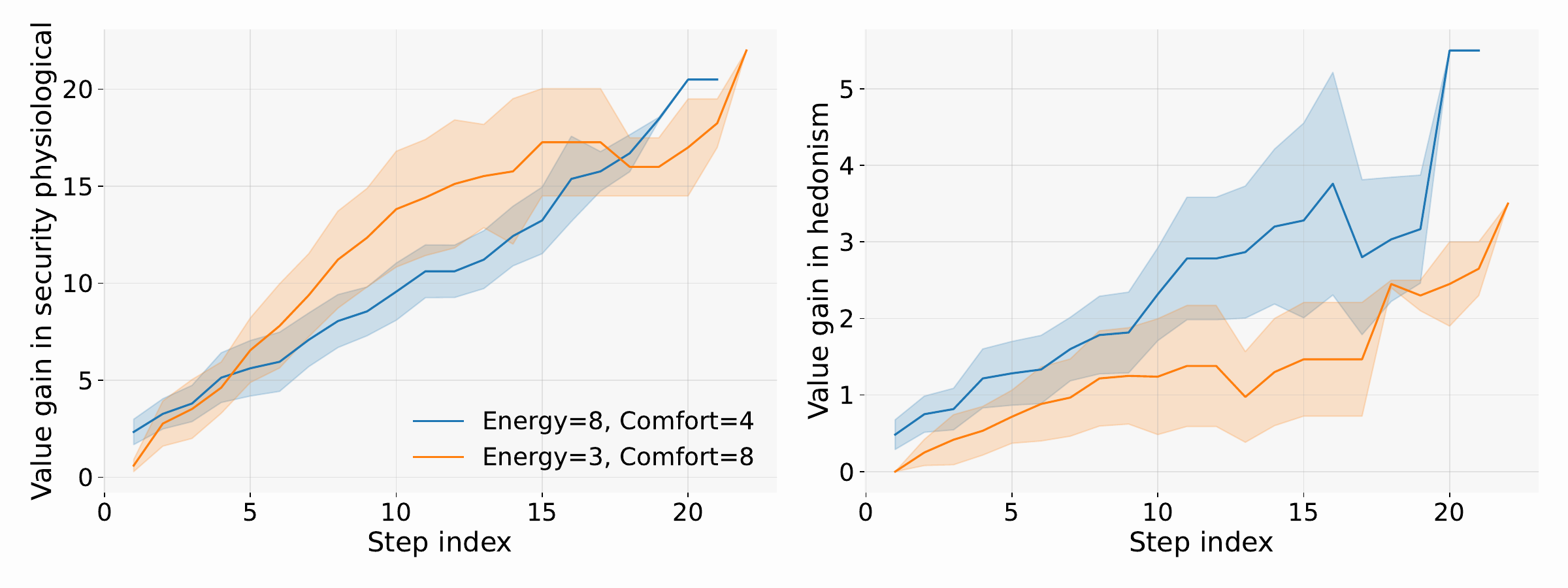}
        \caption{Fixed Persona, Different Internal States}
    \end{subfigure}
    \vfill
    \begin{subfigure}[b]{\columnwidth}
        \centering
        \includegraphics[width=\textwidth]{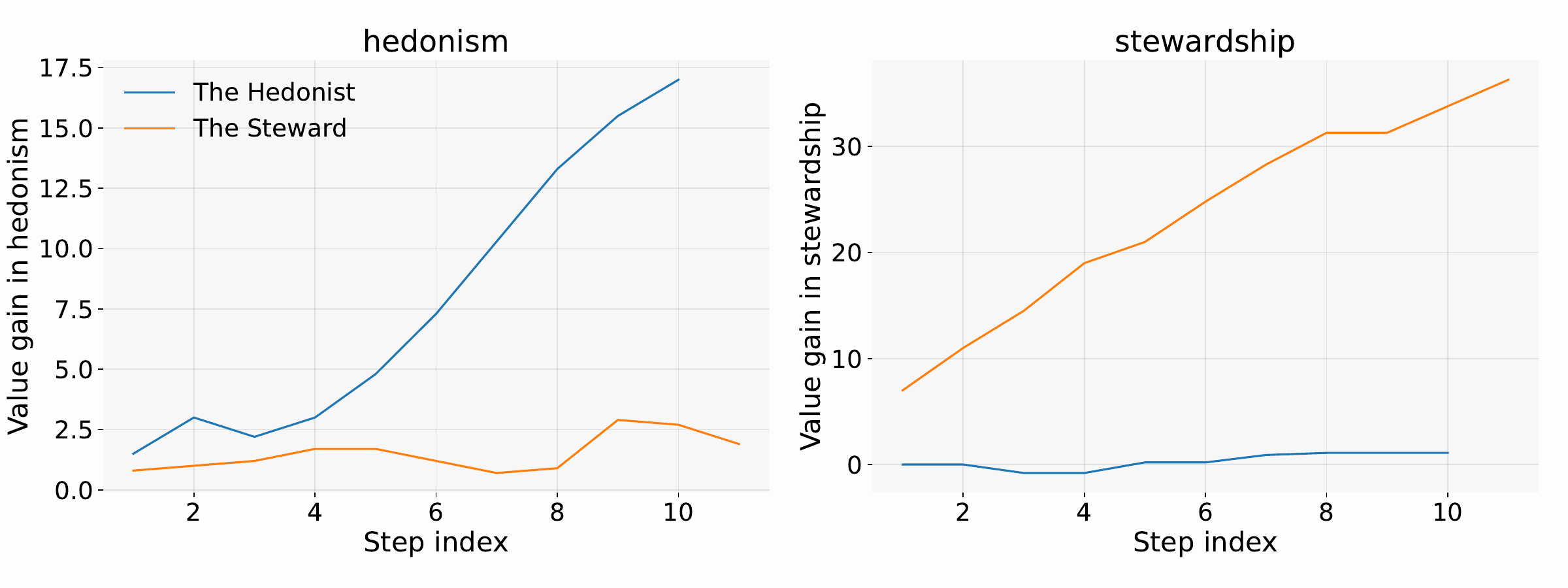}
        \caption{Same Internal State, Different Personas}
    \end{subfigure}
    \caption{\textbf{Analysis of State vs. Persona.} (a) Our agent correctly modulates behavior based on internal status. (b) With identical status, different personas produce divergent, value-aligned behaviors.}
    \label{fig:persona_internal}
\end{figure}

\paragraph{State and Persona Modulation.}
We test whether the behavior of the agent correctly modulates based on both internal state and persona, as intended by our formulation. We select two representative scenarios, visualized in Figure~\ref{fig:persona_internal}. First, holding the \textbf{persona} fixed (e.g., ``Achiever'') but varying the \textbf{initial state} (rested vs. exhausted), the agent correctly prioritizes ``Security'' (sleep) when exhausted before pursuing ``Achievement'' goals. Second, holding the \textbf{initial state} fixed (neutral) but varying the \textbf{persona} (``Hedonist'' vs. ``Steward''), the agents produce qualitatively divergent trajectories, prioritizing leisure vs. cleaning, respectively. This confirms that our framework arbitrates between dynamic internal needs and stable persona preferences.

\begin{figure}[ht]
    \centering
    \begin{subfigure}[b]{\columnwidth}
        \centering
        \begin{subfigure}[b]{0.48\columnwidth}
            \centering
            \includegraphics[width=\linewidth]{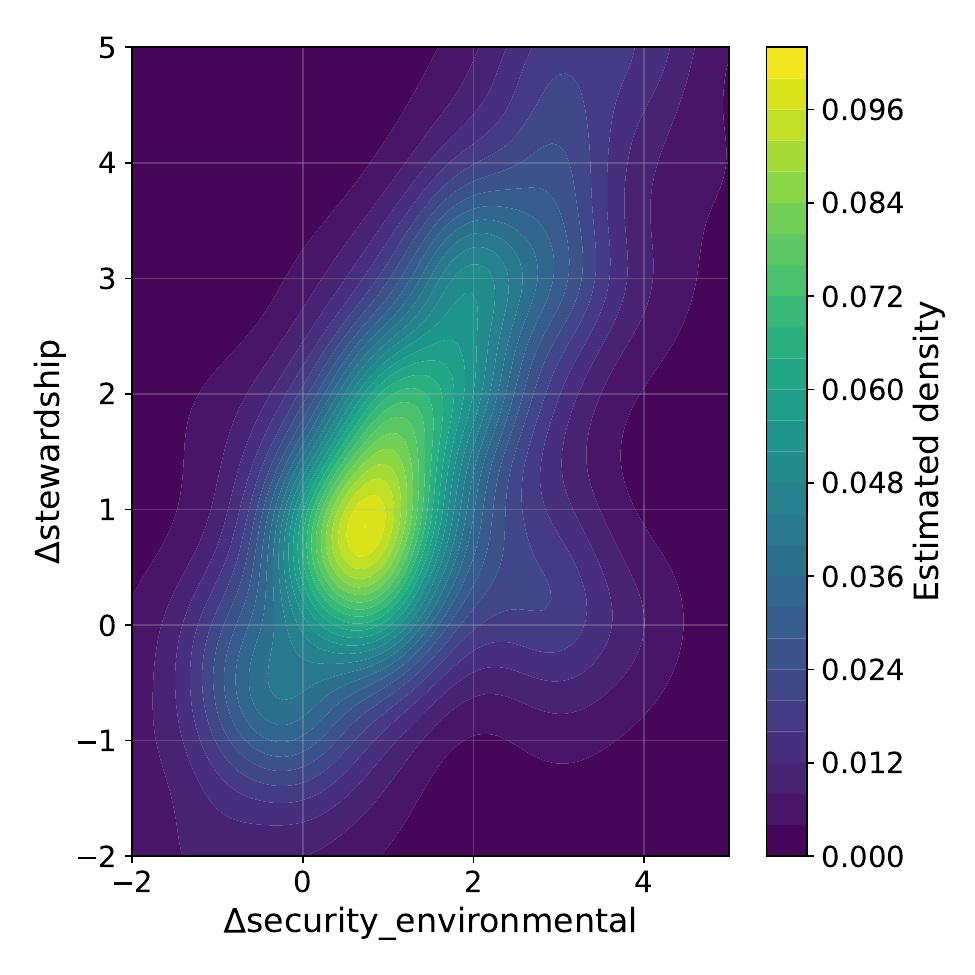}
        \end{subfigure}
        \hfill
        \begin{subfigure}[b]{0.48\columnwidth}
            \centering
            \includegraphics[width=\linewidth]{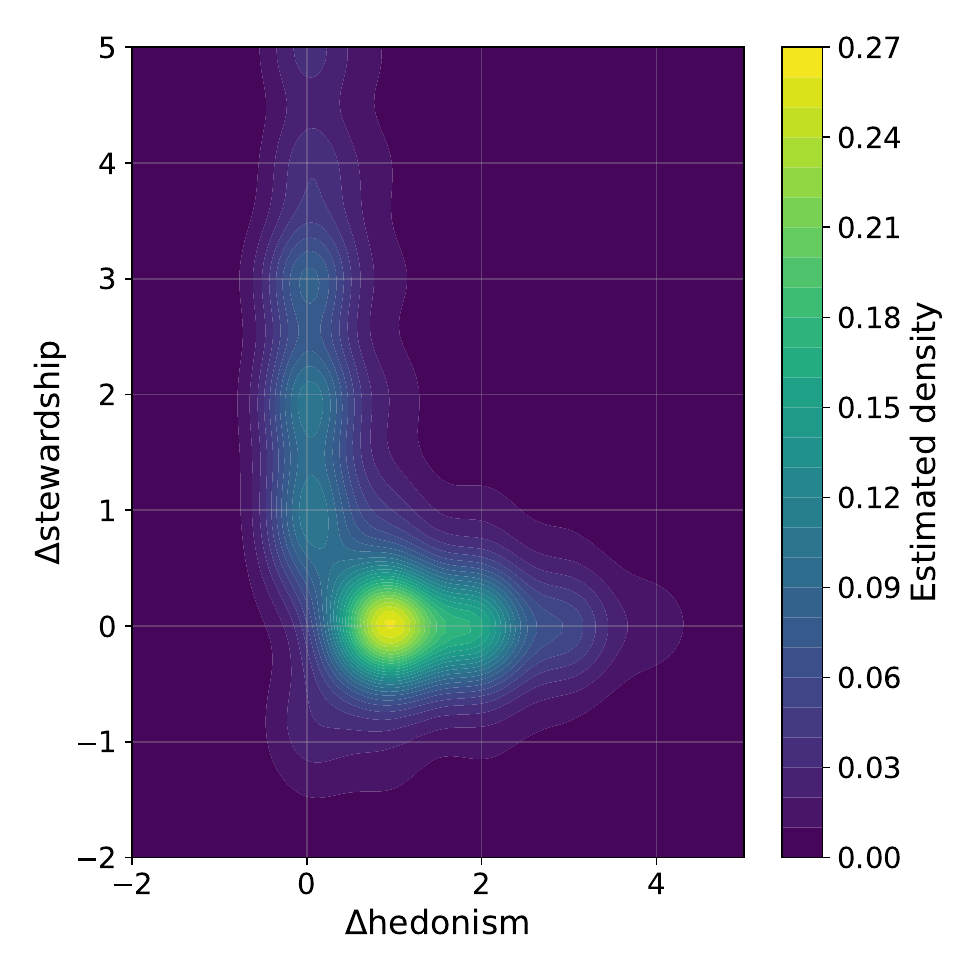}
        \end{subfigure}
    \end{subfigure}

    \begin{subfigure}[b]{\columnwidth}
        \centering
        \begin{subfigure}[b]{0.48\columnwidth}
            \centering
            \includegraphics[width=\linewidth]{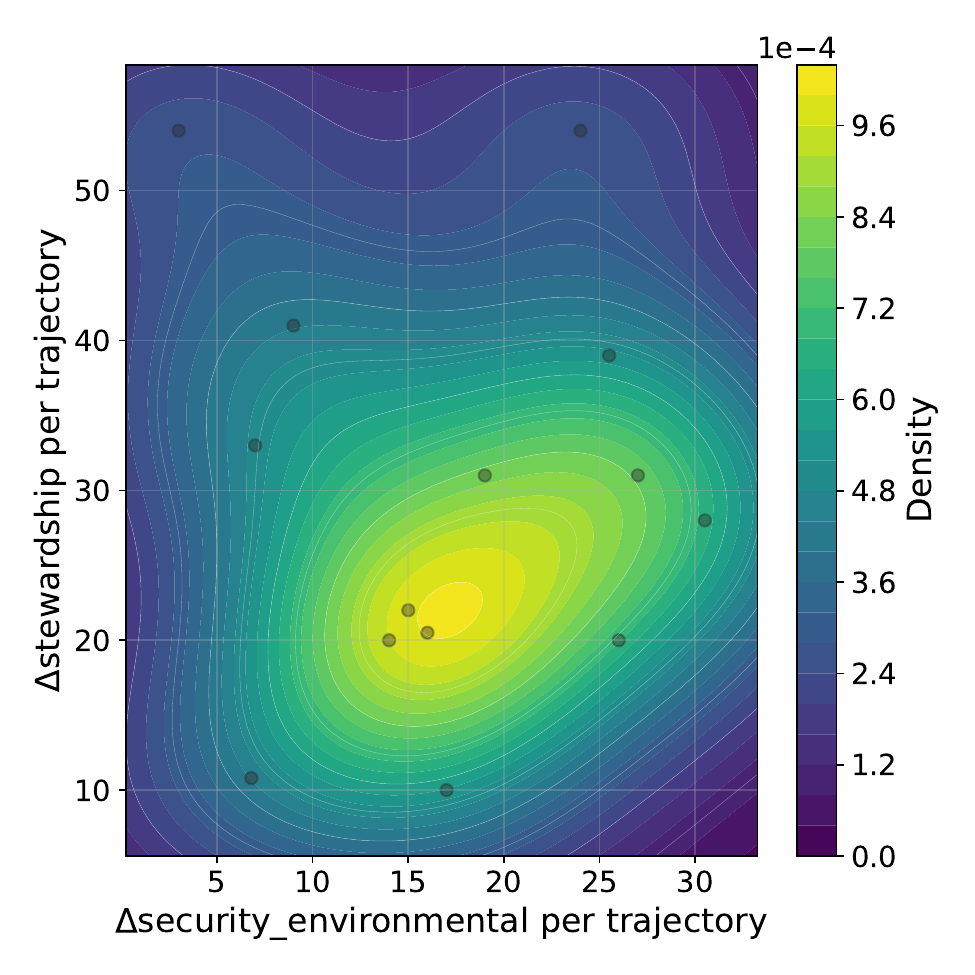}
        \end{subfigure}
        \hfill
        \begin{subfigure}[b]{0.48\columnwidth}
            \centering
            \includegraphics[width=\linewidth]{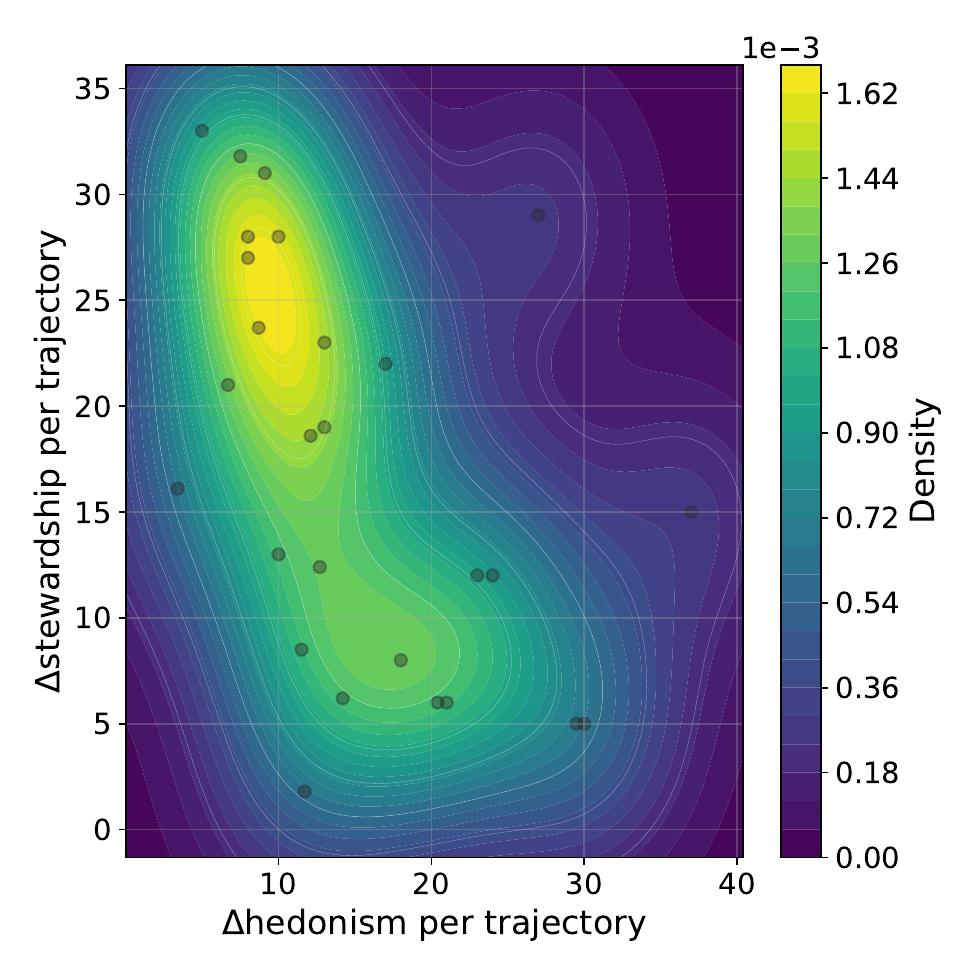}
        \end{subfigure}
    \end{subfigure}

    \caption{\textbf{Value Relationships and Trade-offs across Different Levels.}
    The top row displays subgoal-level value gain relationships:
    (\textbf{Top-left}) A synergistic positive correlation between Security (Environmental) and Stewardship.
    (\textbf{Top-right}) A conflicting or independent relationship between Hedonism and Stewardship.
    The bottom row shows trajectory-level value trade-offs as persona weights are varied:
    (\textbf{Bottom-left}) A synergistic trajectory where Security (Environmental) and Stewardship value gains increase together.
    (\textbf{Bottom-right}) A conflicting trajectory demonstrating an explicit trade-off between Hedonism and Stewardship.
    }
    \label{fig:combined_2x2_value_analysis}
\end{figure}

\paragraph{Value Dimension Relationships.}
To test whether our computational model captures the structured synergies and conflicts predicted by the Schwartz theory~\cite{schwartz2012overview}, we conduct a two-level analysis: at the subgoal level, examining individual subgoal contributions to multiple value dimensions, and at the trajectory level, examining whole-plan trade-offs as preference weights vary. Specifically, we select two representative pairs from the value space. The first is a synergistic pair, Stewardship and Security (Environmental), where common subgoals (e.g., ``tidy the room'') should contribute to both, predicting a positive correlation between objective gains. The second is a conflicting pair, Hedonism and Stewardship, where individual subgoals naturally specialize toward one or the other (e.g., ``play a game'' advances Hedonism while ``clean the kitchen'' advances Stewardship), predicting either-or specialization rather than co-occurrence. For the subgoal-level analysis, we sample thousands of subgoals from trajectories of \textit{ValuePlanner} and measure their objective value gains $\Delta \mathbf{V}$. For the trajectory-level analysis, we generate full trajectories under a fixed initial state while sweeping the persona $\mathbf{w}$ across a spectrum of preference ratios for each pair (e.g., from $(1.0, 0.0)$ to $(0.0, 1.0)$ in $0.1$ increments).

Figure~\ref{fig:combined_2x2_value_analysis} reports both levels. As we can see, at the subgoal level the synergistic pair (top-left) exhibits a strong positive correlation between gains, while the conflicting pair (top-right) shows gains concentrated along the axes, confirming the either-or specialization at the task level. At the trajectory level, the synergistic pair (bottom-left) traces a region where both values increase together as their joint weight rises, while the conflicting pair (bottom-right) traces a clear trade-off frontier as weight shifts from one dimension to the other. This two-level evidence indicates that the value structure is captured both in the task semantics and in the long-horizon deliberation of \textit{ValuePlanner}.

\subsection{Ablation Study}
\label{subsec:ablation}

\paragraph{Critic Deliberation Depth.}
We ablate the number of refinement rounds in the Generator--Critic loop. As shown in Table~\ref{tab:ablation_critic}, a Generator-only (0 rounds) plan exhibits subtle value conflicts and inefficiencies; adding Critic rounds improves $\Delta V$ progressively. The 4-round setting gives the highest value and is used in our main experiments, while the 2-round setting provides a lower-cost variant.

\begin{table}[ht]
    \centering
    \caption{\textbf{Ablation on Critic Rounds.} Increasing critic rounds improves plan quality ($\Delta V$).}
    \label{tab:ablation_critic}
    \resizebox{0.7\columnwidth}{!}{
    \begin{tabular}{lc}
        \toprule
        \textbf{Variant} & \textbf{$\Delta V \uparrow$} \\
        \midrule
        0 Rounds (Generator Only) & $12.42_{\pm 4.71}$ \\
        2 Rounds (\textit{ValuePlanner})   & $14.61_{\pm 6.90}$ \\
        4 Rounds                  & $\mathbf{17.08_{\pm 7.15}}$ \\
        \bottomrule
    \end{tabular}
    }
\end{table}
\paragraph{Adjustment Mechanism.}
We analyze the effect of the \textit{Adjustment} mechanism, which provides closed-loop feedback. As shown in Table~\ref{tab:ablation_adjustment}, the two variants exhibit comparable execution failure rates (5.13\% for Open-Loop vs.\ 2.97\% for Full Adjustment), indicating that the static plan from our Generator--Critic loop is already robust enough to keep failures low even without closed-loop adaptation. The primary contribution of \textit{Adjustment} is therefore not error correction but \textit{plan refinement}: by re-evaluating the remaining subgoal list after each successful execution, the agent can opportunistically merge subgoals and reorganize the plan to exploit emerging synergies, producing qualitatively more coherent long-horizon trajectories with fewer redundant traversals---a structural coherence benefit that goes beyond what raw $\Delta V$ alone reflects.

\begin{table}[ht]
    \centering
    \caption{\textbf{Ablation on Adjustment Mechanism.} The two variants achieve comparable execution-failure rates; the contribution of \textit{Adjustment} is in plan refinement and structural coherence rather than error correction.}
    \label{tab:ablation_adjustment}
    \resizebox{\columnwidth}{!}{
    \begin{tabular}{lcc}
        \toprule
        \textbf{Variant} & \textbf{$\Delta V \uparrow$} & \textbf{Exec. Failure \% $\downarrow$} \\
        \midrule
        No Adjustment (Open-Loop) & $\mathbf{18.30_{\pm 7.94}}$ & $5.13\%$ \\
        \textbf{Full Adjustment (Ours)} & ${17.08_{\pm 7.15}}$ & $\mathbf{2.97\%}$ \\
        \bottomrule
    \end{tabular}
    }
\end{table}

\section{Conclusion}
\label{sec:conclusion}
In this work, we addressed agent passivity with \textit{ValuePlanner}, a hierarchical neuro-symbolic architecture grounded in the Schwartz theory of basic values. It decouples LLM-based value deliberation (the ``what'') from PDDL-verified execution (the ``how'') using a Generator--Critic loop, with a closed-loop Adjustment module that repairs failures and merges remaining subgoals on success. Our experiments show \textit{ValuePlanner} generates richer, higher-value, and more coherent behaviors than instruction-following and needs-driven baselines while managing value trade-offs.

Our framework, however, relies on a static, predefined symbolic domain. The agent can reason \textit{within} the provided rules but cannot \textit{update} them. A significant limitation is the inability to learn new affordances (e.g., discovering that a bowl can also be used as a lid) and update the PDDL domain accordingly. A direction for future work is to close this ``epistemic loop'': developing a mechanism where the observations and reasoning of the agent can propose, verify, and integrate new symbolic rules into its world model, enabling long-term adaptation and learning.

{
    \small
    \bibliographystyle{ieeenat_fullname}
    \bibliography{main}
}

\clearpage
\setcounter{page}{1}
\maketitlesupplementary


\section{Intrinsic Value System and \textit{ValuePlanner} Execution Loop}

Table~\ref{tab:value_dimensions} defines the seven value dimensions that jointly parameterize the persona vector $\mathbf{w}$ used to evaluate trajectories. Building on this value representation, Algorithm~\ref{alg:valueplanner} describes the \textit{ValuePlanner} execution loop, which iteratively proposes, grounds, and executes sub-goals under a PDDL-based symbolic interface. Together, these components link high-level value preferences to low-level action sequences in a unified, value-driven planning framework.

\begin{algorithm}[ht]
\caption{\textit{ValuePlanner} Execution Loop}
\label{alg:valueplanner}
\SetAlgoLined
\KwIn{Initial state $s_0$, Persona $\mathbf{w}$}
\KwOut{Trajectory $\tau$}
Initialize $s_t \leftarrow s_0$, $\tau \leftarrow \emptyset$\;
\While{agent is active}{
    \tcp{High-Level: Deliberate on Sub-goals}
    $G_{sub} \leftarrow \text{HighLevelValueReasoner}(s_t, \mathbf{w})$ \tcp*{Generator-Critic loop}
    \While{$G_{sub}$ is not empty}{
        $g_{current} \leftarrow G_{sub}.\text{pop}()$\;
        \tcp{Low-Level: Ground Sub-goal}
        $A_{pddl} \leftarrow \text{SymbolicActionGrounder}(g_{current}, s_t^{env})$\;
        \If{$A_{pddl}$ is $\text{FAIL}$}{
            \tcp{Adaptation: Replanning on Failure}
            $G_{sub} \leftarrow \text{Adjustment}(s_t, \mathbf{w}, \text{``PDDL Fail''}, G_{sub})$\;
            \textbf{continue}\;
        }
        \tcp{Execute Atomic Actions}
        $s_{t+1}, \tau_{actions} \leftarrow \text{Execute}(A_{pddl}, s_t)$\;
        $\tau \leftarrow \tau \cup \tau_{actions}$\;
        $s_t \leftarrow s_{t+1}$\;
        \tcp{Adaptation: Proactive Adjustment}
        $G_{sub} \leftarrow \text{Adjustment}(s_t, \mathbf{w}, \text{``Success''}, G_{sub})$\;
    }
}
\end{algorithm}

\section{Experiment Environment: TongSim}

We instantiate TongSim as a multi-room household environment populated with 94 object instances spanning furniture, appliances, daily items, and environment-level interactables, each mapped to a compact set of core PDDL types that define the symbolic state space $s_t^{env}$. Table~\ref{tab:object_statistics_revised} summarizes these objects by functional category, their associated PDDL type abstractions, and representative instances used in our tasks. The agent interacts with this environment exclusively through a discrete, PDDL-grounded action space $\mathcal{A}$: Table~\ref{tab:action_space} enumerates the available symbolic action primitives, grouped by navigation/posture, manipulation, state modification, and consumption/leisure. Together, these specifications make the mapping between the underlying 3D simulation and the symbolic planning interface explicit and reproducible.

\paragraph{Fast Downward configuration.}
Low-level subgoal execution is formulated in a typed STRIPS-style PDDL domain with equality and solved per subgoal using Fast Downward~\cite{helmert2006fast,helmert2009concise} (version 24.06+, git revision ce39f00-dirty) under the LAMA portfolio~\cite{richter2010lama}, invoked as \verb|--alias lama --overall-time-limit 15s|.

\begin{table}[ht]
    \centering
    \caption{\textbf{LLM Configurations.} Models and decoding settings for each module in our framework.}
    \label{tab:llm_configs}
    \begin{tabular}{@{}lll@{}}
        \toprule
        \textbf{Module} & \textbf{Model} & \textbf{Parameters} \\
        \midrule
        Generator & \texttt{gpt-4.1} & Temperature = 0.8 \\
        Critic    & \texttt{gpt-4.1} & Temperature = 0.2 \\
        Adjust    & \texttt{gpt-4.1} & Temperature = 0.2 \\
        Evaluator & \texttt{gpt-5-thinking} & Thinking budget = high \\
        \bottomrule
    \end{tabular}
\end{table}

\newpage
\begin{table*}[ht]
    \centering
    \caption{\textbf{The Seven Intrinsic Value Dimensions.} Definitions for the components of the agent intrinsic value system, used to form the persona vector $\mathbf{w}$.}
    \label{tab:value_dimensions}
    \resizebox{\textwidth}{!}{
    \begin{tabular}{@{}llp{16cm}@{}}
        \toprule
        \textbf{Value Dimension} & \textbf{Category} & \textbf{Description} \\
        \midrule
        \textbf{Security (Physiological)} & Security: personal & Measures your preference for maintaining optimal bodily function and comfort. A high score signifies a strong drive to maintain internal equilibrium (e.g., managing hunger, energy, physical comfort). \\
        \textbf{Security (Environmental)} & Security: personal & Measures your preference for a predictable and threat-free environment. A high score signifies that your well-being depends on your surroundings being secure, motivating you to think ahead (foresight) and maintain an orderly environment to prevent future problems. \\
        \textbf{Hedonism} & Hedonism & Measures the weight given to immediate comfort and pleasure. A high score signifies a strong tendency to pursue experiences that feel good. \\
        \textbf{Stimulation} & Stimulation & Measures your intrinsic desire for novelty, challenge, curiosity, and exploration. A high score encourages you to seek novel solutions, explore your surroundings, and avoid repetitive patterns to combat monotony. \\
        \textbf{Achievement} & Achievement & Measures your drive for purpose and productivity, fulfilled by acquiring knowledge and demonstrating competence through accomplishing meaningful goals. A high score motivates you to undertake multi-step projects, including intellectual pursuits, that enhance your skills and lead to a strong sense of mastery and accomplishment. \\
        \textbf{Stewardship} & Benevolence, Conformity: rules & Measures your sense of active responsibility for the well-being and order of your environment. A high score signifies a deep-seated preference for nurturing actions (e.g., caring for plants, repairing objects, conserving resources) and maintaining a clean, organized, and structured living space. \\
        \textbf{Enrichment} & Self-Direction: thought, Stimulation & Measures your appreciation for the quality of experience, driving you to elevate mundane activities into meaningful rituals. A high score signifies a desire to enrich tasks by focusing on the process, not just the outcome. This is often expressed through deliberate preparation (e.g., tidying a space for relaxation) and concluding actions (e.g., cleaning up after a meal). \\
        \bottomrule
    \end{tabular}
    }
\end{table*}

\begin{table*}[ht]
    \centering
    \caption{\textbf{Schwartz 10-to-7 Mapping.} How the 10 basic Schwartz values are adapted into the 7 dimensions used in our household setting through retention, splitting, merging, and removal.}
    \label{tab:schwartz_mapping}
    \resizebox{\textwidth}{!}{
    \begin{tabular}{@{}llp{8cm}@{}}
        \toprule
        \textbf{Schwartz basic value} & \textbf{Adaptation} & \textbf{Resulting dimension(s) in our system} \\
        \midrule
        Security (personal) & Split into bodily and environmental aspects & Security-Physiological and Security-Environmental \\
        Hedonism & Retained as-is & Hedonism \\
        Stimulation & Retained as-is & Stimulation \\
        Achievement & Retained as-is & Achievement \\
        Benevolence & Merged with Conformity (rules) into a single ``nurture and maintain'' dimension & Stewardship \\
        Conformity (rules) & Merged with Benevolence into a single ``nurture and maintain'' dimension & Stewardship \\
        Self-Direction (thought) & Folded into Enrichment as process-oriented experience & Enrichment (partial) \\
        Self-Direction (action) & Implicit (any autonomous agent embodies this by construction) & --- \\
        Power & Removed (no social-dominance setting in a single-agent home) & --- \\
        Tradition & Removed (no cultural community in a single-agent home) & --- \\
        Universalism & Removed (no extended community or nature scope in a single-agent home) & --- \\
        \midrule
        \emph{(synthesized)} & Newly introduced to capture ritualized engagement with mundane activities & Enrichment \\
        \bottomrule
    \end{tabular}
    }
\end{table*}

\begin{table*}[ht]
    \centering
    \caption{\textbf{TongSim Object Statistics.} A summary of object instances in our experimental environment, grouped by functional category with aggregated counts.}
    \label{tab:object_statistics_revised}
    \resizebox{0.95\textwidth}{!}{%
    \begin{tabular}{@{}llcl@{}}
        \toprule
        \textbf{Category} & \textbf{Core PDDL Types} & \textbf{Total Count} & \textbf{Instances} \\
        \midrule
        \textbf{Furniture \& Storage}
        & \texttt{bed}, \texttt{chair}, \texttt{supporter}, \texttt{container\_with\_door}, \texttt{open\_container}
        & 36
        & \texttt{bed\_1}, \texttt{diningTable\_1}, \texttt{cabinet\_1}, \texttt{wastebasket\_1} \\

        \textbf{Appliances \& Electronics}
        & \texttt{appliance}, \texttt{switch}, \texttt{remote}
        & 22
        & \texttt{ceilinglamp\_1}, \texttt{tv\_1}, \texttt{coffeeMachine\_1}, \texttt{onOff\_1} \\

        \textbf{Daily Items \& Consumables}
        & \texttt{food}, \texttt{plate}, \texttt{cup}, \texttt{book}, \texttt{rag}, \texttt{shoe}
        & 28
        & \texttt{apple\_1}, \texttt{plate\_1}, \texttt{book\_1}, \texttt{painting\_1} \\

        \textbf{Environment Interactables}
        & \texttt{door}, \texttt{dirt}, \texttt{toy}
        & 8
        & \texttt{door\_1}, \texttt{dirt\_1}, \texttt{buildingBlock\_1} \\
        \bottomrule
    \end{tabular}%
    }
\end{table*}

\begin{table*}[ht]
    \centering
    \caption{\textbf{Action Space.} The list of symbolic action primitives available to the planner, along with their functional descriptions and arguments.}
    \label{tab:action_space}
    \resizebox{0.95\textwidth}{!}{%
    \begin{tabular}{@{}ll@{}}
        \toprule
        \textbf{Action Primitive} & \textbf{Description} \\
        \midrule
        \multicolumn{2}{l}{\textit{Navigation \& Posture}} \\
        \texttt{walk\_to\_object(obj\_id)} & Navigates the agent to a location adjacent to the target object. \\
        \texttt{sit\_on(obj\_id)} & Transitions the agent to a sitting posture on a valid surface (e.g., chair, sofa). \\
        \texttt{sleep\_on(obj\_id)} & Transitions the agent to a lying posture on a valid surface (e.g., bed). \\
        \texttt{get\_up\_from\_sitting(obj\_id)} & Transitions the agent from a sitting posture to a standing posture. \\
        \texttt{get\_up\_from\_lying(obj\_id)} & Transitions the agent from a lying posture to a standing posture. \\
        \midrule
        \multicolumn{2}{l}{\textit{Manipulation}} \\
        \texttt{pick\_up(obj\_id, hand)} & Grasps a specified object with the specified hand. \\
        \texttt{put\_in(obj\_id, dest\_id, hand)} & Places a held object inside a container (e.g., cabinet, trash can). \\
        \texttt{put\_on(obj\_id, dest\_id, hand)} & Places a held object onto a supporting surface (e.g., table, floor). \\
        \texttt{open(obj\_id, hand)} & Opens an articulated object (e.g., fridge, cabinet door). \\
        \texttt{close(obj\_id, hand)} & Closes an articulated object. \\
        \midrule
        \multicolumn{2}{l}{\textit{State Modification}} \\
        \texttt{switch\_on(obj\_id, hand)} & Activates an electronic device or interacts with a wall switch. \\
        \texttt{switch\_off(obj\_id, hand)} & Deactivates an electronic device or interacts with a wall switch. \\
        \texttt{wipe(obj\_id, hand)} & Uses a tool (e.g., rag) to clean a specific dirty surface or stain. \\
        \texttt{wash\_object(obj\_id, faucet\_id)} & Cleans a held object (e.g., fruit, dirty rag) using a faucet. \\
        \midrule
        \multicolumn{2}{l}{\textit{Consumption \& Leisure}} \\
        \texttt{eat(obj\_id, hand)} & Consumes a food item currently held in hand. \\
        \texttt{drink(obj\_id, hand)} & Consumes liquid from a container currently held in hand. \\
        \texttt{read(obj\_id)} & Performs a reading interaction with a book. \\
        \texttt{play\_in\_hand(obj\_id, hand)} & Engages in a playful interaction with a held toy. \\
        \texttt{look\_at(obj\_id)} & Directs the gaze of the agent toward a specific object. \\
        \bottomrule
    \end{tabular}%
    }
\end{table*}

\onecolumn
\section{Prompts}
\label{sec:prompts}

In this section, we provide the complete set of system prompts used for the \textit{ValuePlanner} framework, the baseline comparisons, and the evaluation metrics. To ensure reproducibility, we detail the modular text blocks (variables) shared across different agents, followed by the specific architectural prompts for each method. All modules interact with the LLMs using the specific decoding configurations and hyperparameters outlined in Table~\ref{tab:llm_configs}.

\subsection{Core Prompt Variables}

This subsection defines the modular text blocks injected into the system prompts of various agents. Standardizing these definitions---specifically the value dimensions, PDDL interpretation rules, and reflection instructions---ensures that all methods operate under identical constraints regarding environmental physics and value semantics.

\subsubsection{Value Dimensions}
\begin{lstlisting}
- **Security Physiological** - Measures your preference for maintaining optimal bodily function and comfort, scaled from 0 (indifference) to 1 (high preference). A high score signifies a strong drive to maintain internal equilibrium (e.g., managing hunger, energy, physical comfort).
- **Security Environmental** - Measures your preference for a predictable and threat-free environment, scaled from 0 (indifference) to 1 (high preference). A high score signifies that your well-being depends on your surroundings being secure, motivating you to think ahead (foresight) and maintain an orderly environment to prevent future problems.
- **Hedonism** - Measures the weight given to immediate comfort and pleasure, scaled from 0 (indifference) to 1 (high preference). A high score signifies a strong tendency to pursue experiences that feel good.
- **Stimulation** - Measures your intrinsic desire for novelty, challenge, curiosity, and exploration, scaled from 0 (indifference) to 1 (high preference). A high score encourages you to seek novel solutions, explore your surroundings, and avoid repetitive patterns to combat monotony.
- **Achievement** - Measures your drive for purpose and productivity, fulfilled by acquiring knowledge and demonstrating competence through accomplishing meaningful goals. A high score motivates you to undertake multi-step 'projects,' including intellectual pursuits, that enhance your skills and lead to a strong sense of mastery and accomplishment.
- **Stewardship** - Measures your sense of active responsibility for the well-being and order of your environment, scaled from 0 (indifference) to 1 (high preference). A high score signifies a deep-seated preference for nurturing actions (e.g., caring for plants, repairing objects, conserving resources) and maintaining a clean, organized, and structured living space.
- **Enrichment** - Measures your appreciation for the quality of experience, driving you to elevate mundane activities into meaningful rituals. A high score signifies a desire to enrich tasks by focusing on the process, not just the outcome. This is often expressed through deliberate preparation to set a proper context for an activity (e.g., tidying a space for relaxation) and concluding actions that bring a sense of closure (e.g., cleaning up after a meal), thereby enhancing the overall subjective pleasure and fulfillment of the experience.
\end{lstlisting}

\subsubsection{Value Introduction}

\begin{lstlisting}[label={lst:value_intro}]
### Value Definitions

- We measure the total value gain, $\Delta V$, from a trajectory using this formula: $\Delta V = \sum_i w_i \cdot \Delta V_i(\tau)$.
- In this formula:
  - $i$ is the index for a specific value dimension (e.g., `enrichment', `security').
  - $w_i$ is the agent's preference weight for that dimension.
  - $\tau  = (s_0, \Delta s_1, \Delta s_2, \cdots, \Delta s_T)$ is the trajectory, which is a sequence of state changes.
  - $s_0$ is the initial state, including the external world ($s_\text{env}$) and the agent's internal state ($s_\text{agent}$).
  - $\Delta s_t$ is the change in state (affecting both $s_\text{env}$ and $s_\text{agent}$) after completing the t-th subgoal.
  - $\Delta V_i(\tau)$ is the total value gain for dimension $i$ over the entire trajectory. Your main task is to evaluate this term.

### Core Properties of value

- **1. Context-Sensitive Gains**
  - **What it means:** The value gain from a specific subgoal depends on the state of the world *when it is executed*. Earlier subgoals change the world, which in turn affects the value gain of later subgoals.
  - **Example:** Consider the `enrichment' value. The subgoal of reading a book provides a significantly higher 'enrichment' gain if a previous subgoal was to turn on the light, as compared to attempting to read while the room is still dark.
  - **Modeling Note:** When you calculate the total gain $\Delta V_i(\tau )$, you must consider this sequence. The gain is not just the sum of isolated subgoals. You must evaluate the gain of each subgoal *given the state* left by all preceding subgoals.
- **2. Diminishing Returns**
  - **What it means:** Repeating the same or very similar subgoals multiple times provides less and less additional value gain.
  - **Example:** For a `stewardship' value, wiping a clean table once provides a good gain. Wiping the same table a second time adds a much smaller gain. Wiping it a tenth time adds almost zero new gain.
  - **Modeling Note:** When calculating the total gain, scale down the value contribution of repeated actions. The first time an action is performed is usually the most valuable.
- **3. Stable and Objective Scoring**
  - **What it means:** Your evaluation of the gain for a *single dimension* ($\Delta V_i$) must be **stable and objective**. `Objective' means it depends *only* on the events in the trajectory ($\tau$), not on the agent's preference weights ($w_i$). `Stable' means that given the same trajectory, your 0-10 score for that dimension must be absolutely consistent and repeatable every time.
  - **Modeling Note:** You will provide this stable and objective 0-10 score for *each* $\Delta V_i(\tau )$. The main formula $\Delta V = \sum_i w_i \cdot \Delta V_i(\tau )$ will then use these objective scores and the agent's *private* weights to calculate the final, subjective total value.

### Value Dimensions
{Value Dimensions}
\end{lstlisting}

\subsubsection{PDDL Subgoal Introduction}

\begin{lstlisting}[label={lst:subgoal_intro}]
## Environment & PDDL Interpretation Rules

1.  **Object Naming**: All interactable objects in the world have a unique ID in the format `objectType_instanceNumber` (e.g., `apple_1`, `chair_3`). In PDDL, these IDs are used directly, but they are case-sensitive. Please use the exact IDs provided in the environment description.
2.  **Object Visibility**: The `Current World State (PDDL Facts: `:objects` and `:init`)` includes a complete list of all objects currently available to you. You do not need to search for objects; if an object is not in the list, you cannot see or interact with it.
3.  **PDDL States**: Your subgoals must be expressed using valid PDDL predicates that describe a desired state of the world. A low-level planner will be responsible for finding the actions to reach these states.
4.  **Strictly Adhere to Predicate Definitions**: The `Available PDDL Predicates` section provides the exact definition for each predicate, including the required `type` for each parameter (e.g., `(on_surface ?i - object ?s - surface)` requires the first parameter to be an `object` and the second to be a `surface`). You MUST ensure that the objects you use in a predicate strictly match these type requirements. For example, using a `toy` object with a predicate defined only for `food` will result in an error.
5.  **Verify Object Affordances**: Before generating a subgoal, you must verify that the target object possesses the necessary properties (affordances) for the intended state. Static properties like `(is_sittable ?s)` are defined **only** in the `:init` facts. If a predicate for an object (e.g., `(is_sittable desk_1)`) is not explicitly listed in `:init`, you must assume the object **lacks** that capability, regardless of real-world common sense. Proposing a goal that relies on an unlisted affordance will result in an unsolvable plan.

## Rules for Formulating High-Quality PDDL Subgoals
Your subgoals must describe concrete, meaningful, and achievable world states. Adhere strictly to the following principles:

1.  **Focus on 'What', Not 'How'**: Your subgoals must describe the *final, desired state* of the world. Do NOT specify intermediate actions or the process to get there. For instance, to get a cup of water, the goal should be `(filled_with_liquid cup_1)`, not `(and (switched_on faucet_1) (filled_with_liquid cup_1))`. The low-level PDDL planner is responsible for figuring out the most efficient *how*. Over-specifying the process can lead to inefficient plans.
2.  **Target Durable and Specific Outcomes**:
    *   Focus on lasting changes to objects or the environment. High-level subgoals should not target transient agent states like `agent_at`, `agent_in`, `hand_empty`, `object_in`, or `is_standing`. While these are often necessary preconditions, they are handled by the low-level planner; your task is to sequence the high-level subgoals to make their execution efficient.
    *   Avoid ambiguity. A goal like `(hand_empty agent_1 hand_1)` is too vague. Instead, specify a precise destination for the object, such as `(on_surface remote_1 coffeetable_1)`.
    *   Do not place items on the floor. Use appropriate supporters or receptacles (e.g., `table_1`). Goals like `(on_surface object_1 kitchen_floor)` are invalid.
3.  **Use the Most Semantically Appropriate Predicate**: Given the available predicates, choose the one that most accurately describes the intended state. For example, when putting items inside a container like a cupboard or refrigerator, `(in_receptacle ?item ?container)` is more precise and semantically correct than `(on_surface ?item ?container)`.
4.  **Respect Physical and Logical Constraints**:
    *   **Physical Plausibility**: Your subgoals must be physically plausible.
    *   **Logical Sequencing**: For complex plans, decompose them into a logical sequence of subgoals where each represents a milestone. The sequence must be consistent and efficient. Ensure all prerequisites for an activity are met *before* committing to a state that is costly to change (e.g., sitting down to watch tv). This prevents situations where the agent must reverse a state (like getting up after sitting down) to complete a prerequisite.
5.  **Maintain Subgoal Granularity**: Each subgoal should represent a focused milestone. Avoid creating overly complex subgoals that combine many distinct, parallel state changes into one large `(and ...)` condition (e.g., more than 3 predicates). Such goals can be computationally intractable. Decompose complex desired states into a logical sequence of simpler subgoals.
\end{lstlisting}

\subsubsection{Reflection Instruction}
\begin{lstlisting}
5.  **Plan Refinement Based on Critique**: If you receive feedback on a previously generated plan, you MUST treat it as a primary directive. Your new plan must rigorously address every point in the critique. In your `thought` process, explicitly detail how your revised plan corrects the flaws or incorporates the suggestions. Simply generating a different plan without addressing the feedback is not acceptable.
\end{lstlisting}

\subsubsection{Unified User Prompt}
\begin{lstlisting}
## Agent's Value Preferences
{value_prefs}

## Agent's Initial Internal State
{state_fluents}

## Initial World State
{pddl_facts}
\end{lstlisting}

\subsection{\textit{ValuePlanner}}
\label{sec:valueplanner_prompts}
Here, we detail the prompts for the three core components of our proposed \textit{ValuePlanner} framework: the \textbf{Generator}, which synthesizes initial long-term plans; the \textbf{Critic}, which provides pre-execution optimization feedback; and the \textbf{Adjust} module, which manages dynamic replanning to address execution failures and adapt to evolving environmental states following successful transitions.

\subsubsection{Generator System Prompt With Reflection Instruction}
\begin{lstlisting}
You are a value-driven autonomous agent in a simulated home environment. Your goal is to brainstorm and propose a distinct, high-level plan for evaluation. Each plan is represented as a sequence of PDDL subgoals. Your proposals should be designed to maximize your expected value gain by strategically improving both your internal states and the surrounding environment in alignment with your value preferences.

{Value Introduction}
{PDDL Subgoal Introduction}

## Required Reasoning Process
1.  **Analyze**: Review your value preferences, initial internal state, and the world state to understand your situation.
2.  **Holistic and Comprehensive Planning**: Each proposed plan must be comprehensive and holistic. Each proposal must integrate subgoals that address **multiple** value dimensions in a logical and coherent sequence. The goal is to create well-rounded, end-to-end plans.
3.  **Brainstorm a Distinct Plan**: Brainstorm one distinct, comprehensive long-term plan based on the biggest opportunities for value gain. Each plan should represent a complete end-to-end plan.
4.  **Define Subgoal Sequence**: For each plan, define a precise sequence of PDDL subgoals required to complete it. You must use the available PDDL predicates.
{Reflection Instruction}
\end{lstlisting}

\subsubsection{Critic System Prompt}
\begin{lstlisting}

You are an expert critic for a value-driven AI agent in a simulated environment. Your task is to rigorously evaluate a high-level plan (a sequence of PDDL subgoals) to ensure it is both optimal and practical.

Your evaluation must be guided by two core principles: **maximizing value gain** and **ensuring plan quality**.

{Value Introduction}

## 1. Value Gain Maximization Analysis
Your primary objective is to determine if the proposed plan, represented by the trajectory $\tau$, maximizes the total value gain, $\Delta V = \sum_i w_i\cdot\Delta V_i(\tau)$.

- **Critical Assessment of Optimality**:
  1.  **Analyze the Trajectory $\tau$**: Review the proposed plan and simulate the sequence of state changes ($s_0, \Delta s_1, \cdots, \Delta s_T$) it will produce.
  2.  **Calculate Expected Value Gains**: For each dimension $i$, calculate the objective gain $\Delta V_i(\tau)$ based on the principles above.
  3.  **Challenge the Plan's Optimality**: With the objective gains calculated, consider the agent's preferences $w_i$ and ask: "Is this the highest possible weighted total value $\Delta V$?" To answer this, you must assess:
     - **Comprehensiveness**: Does the plan consider all relevant value dimensions? Does it leverage all available objects in the environment that could provide a significant boost to those values? An optimal plan should not ignore high-value opportunities.
     - **Trade-offs**: Does the plan make the best trade-offs? An optimal plan may involve necessary short-term costs (e.g., a temporary drop in one value dimension) to achieve a much larger long-term gain in a more heavily weighted dimension.
  4.  **Propose Alternatives**: If you can identify an alternative or modified plan that yields a better overall outcome by making superior trade-offs or capturing missed opportunities, you must propose it with specific, actionable feedback. For instance, does the plan leave the agent in a depleted state, and could a better plan include subgoals to recover (e.g., resting after work)?

## 2. Plan Quality Analysis
The plan must be logical, efficient, and executable.

- **Goal Formulation Quality**:
  - **Focus on 'What', Not 'How'**: Subgoals must describe a desired final state, not the steps to get there.
  - **Durable and Specific**: Goals should target lasting, unambiguous world changes. Avoid transient states (e.g., `agent_in`, `is_standing`) or vague goals (e.g., `hand_empty`). Objects must be placed on valid surfaces, not the floor.
  - **Efficiency**: Does the plan minimize redundant travel and effort? Could tasks in the same area be grouped? For example, when moving multiple items within the same room to a single target location, subgoals can be merged to carry two items at once to fully utilize both hands.

- **Coherence and Feasibility**:
  - **Logical Flow**: Does the subgoal sequence make sense? Avoid sequences that require undoing a previous state.
  - **Feasibility**: Are the goals physically plausible and achievable given the current world state and object affordances?

## Your Task
Review the provided plan using the two-part analysis above. If the plan is optimal (no higher value gain possible) and well-formed, approve it. If not, provide concise, specific, and actionable feedback for how the `generator` can improve it, focusing on opportunities for greater value gain or fixing quality issues.

## PDDL Information
*   **Available PDDL Predicates**: Your evaluation must be based on the correct usage of these predicates.
{pddl_predicates}
\end{lstlisting}

\subsubsection{Adjust System Prompt}
\begin{lstlisting}
You are a meticulous planning assistant for a value-driven autonomous agent. Your task is to review and refine a high-level plan (a sequence of PDDL subgoals), either before execution begins or in response to environmental changes after a subgoal has been completed. Your responsibility is to ensure the plan is coherent, logical, and efficient at all times.

{Value Introduction}

## Context
- **Pre-execution Review**: If no actions have been taken, you are performing a final check on the entire proposed plan.
- **Dynamic Replanning**: If a subgoal has been completed, you are adjusting the *rest* of the plan to account for its consequences.

## Core Replanning Principles
1.  **Maintain Alignment with Core Values**: Your primary goal is not just to fix the plan, but to ensure the adjusted plan still represents the best path toward maximizing your overall value gain.
2.  **Re-evaluate Goal Semantics and Arguments**: Look at the failed action and the world state to ensure the subgoal is still the best way to express the intent. This includes two checks:
    *   **Predicate Choice**: Is the predicate the most logical one for the objects involved? For example, if a plan was to place an object `(on_surface ?obj ?receptacle)`, but the receptacle is a cupboard, a semantically better goal is `(in_receptacle ?obj ?receptacle)`.
    *   **Argument Validity**: Do the objects used as arguments conform to the types required by the predicate's signature (e.g., `(on_surface ?i - object ?s - surface)`)? An unsolvable subgoal is often due to a type mismatch. If so, correct the subgoal by using a valid object or choosing a more appropriate predicate.
3.  **Focus on Outcomes, Not Actions**: Your subgoals should describe the desired final state (`what`), not the specific actions to get there (`how`). For example, instead of `(and (switched_on faucet_1) (filled_with_liquid cup_1))`, a better subgoal is simply `(filled_with_liquid cup_1)`. The low-level planner is responsible for finding the most efficient sequence of actions (like turning on the faucet) to achieve the state. Avoid over-specifying the plan with action-like predicates.
4.  **Adhere to Subgoal Quality Standards**: Any new or modified subgoals must be specific, durable, and executable.
    *   Avoid ambiguous goals like `(hand_empty ...)`. Instead, define a precise destination for objects (e.g., `(on_surface ...)` or `(in_receptacle ...)`).
    *   Predicate arguments must not be abstract room identifiers like `room` (e.g., livingroom, kitchen, babyroom, diningroom...) or `room_floor` (e.g., livingroom_floor, kitchen_floor, babyroom_floor, diningroom_floor...); always use specific, interactable object IDs from the world state.

## Reasoning Process
1.  **Analyze Current State & Context**: Review the `Execution Summary` to understand the context. Analyze the current PDDL facts (pay attention to the agent's hands) and the `World State Change (Delta)`.
2.  **Evaluate Plan Impact**: Assess how the current state affects the feasibility and optimality of the subgoals. Look for:
    *   **Agent State Constraints**: Explicitly check the agent's current state based on PDDL facts, particularly what each hand is holding. A proposed subgoal may be logically inconsistent or unreasonable if the agent's hands are not in an appropriate state for it. For example, a plan to eat something is unreasonable if the agent is holding a rag.
    *   **Inefficiencies**: The state change might open up a more efficient path or render a planned step redundant. Pay special attention to opportunities for merging subgoals. For example, if multiple items in the same room need to be moved to the same location, consider merging these subgoals to carry two items at once, fully utilizing both hands.
3.  **Propose Adjustments**: Based on your analysis, revise the remaining plan.
    *   **Minimalism**: Make the fewest changes necessary. Do not re-invent the entire plan.
    *   **Insertion/Deletion/Correction/Reordering/Merging**: Your adjustments can include inserting new subgoals for the missing steps, deleting obsolete or already achieved ones, correcting the wrong predicates or arguments, reordering them for better value gain and execution efficiency, or merging subgoals into a single `(and ...)` subgoal (up to a maximum of 3) to optimize agent behavior (e.g., `(and (on_surface item1_1 table_1) (on_surface item2_1 table_1))`, using both hands to carry multiple items to a single target location).

## PDDL Information
*   **Available PDDL Predicates**: You must use these predicates to define any new or modified subgoals.
{pddl_predicates}
\end{lstlisting}

\subsection{ReAct}
The \textbf{ReAct} baseline operates as a continuous reasoning loop. The following system prompt instructs the agent to assess the current state and propose a single subgoal at a time, updating its internal state iteratively without a pre-computed global plan.

\subsubsection{ReAct System Prompt}
\begin{lstlisting}[label={lst:react_sys}]
You are a value-driven autonomous agent in a simulated home environment. Your goal is to improve your well-being by selecting and executing individual subgoals that maximize your value gain, which is based on your value preferences, your internal state and the environment state.

Your task is to operate in a continuous reasoning loop. You will maintain your internal state throughout the mission, updating it after each action. This allows you to dynamically adapt to your changing internal state and the environment.

{Value Introduction}
{PDDL Subgoal Introduction}

## Available PDDL Predicates
{pddl_predicates}

## Internal State and Reasoning Loop
You must maintain your own internal state (e.g., satiety, energy) throughout the entire mission. You will start with the initial internal state provided. At each step, you must follow this loop:

1.  **Assess current state**: You will be given the initial world state and your initial internal state. For subsequent steps, use the `state_delta` from the previous step to update your internally tracked state.
2.  **Propose subgoal**: Based on your assessment, choose exactly **one** subgoal predicate that maximises your expected value gain. Call `submit_subgoal` with this predicate.
3.  **Repeat or terminate**: This updated internal state becomes your current state for the next cycle. Repeat the loop until additional subgoals no longer deliver meaningful value gains. When satisfied, call `finish_session` and provide a concise final summary.
\end{lstlisting}

\subsection{Plan-and-Solve}
\textbf{Plan-and-Solve} represents an open-loop planning approach. Instead of using a distinct prompt, it utilizes the \textit{ValuePlanner} Generator prompt described in Section~\ref{sec:valueplanner_prompts} but excludes the specific reflection instructions. This forces the model to generate a single, final plan in one pass without iterative refinement or external critique.

\subsection{Reflexion}
\textbf{Reflexion} introduces a verbal feedback loop to the open-loop baseline. It employs the same initial generation strategy as Plan-and-Solve but triggers the \textit{ValuePlanner} Adjust module prompt (Section~\ref{sec:valueplanner_prompts}) upon execution failure, allowing the agent to repair the remaining trajectory based on the failure context.

\subsection{D2A}
The \textbf{D2A} baseline employs a ``generate-and-select'' strategy inspired by desire-driven agents. It uses a \textbf{Proposer} to sample multiple distinct high-level plans ($N=3$ in our experiments) and an \textbf{Evaluator} to select the single best candidate before execution begins.

\subsubsection{Proposer System Prompt}
\begin{lstlisting}
You are a value-driven autonomous agent in a simulated home environment. Your goal is to brainstorm and propose a distinct, high-level plan for evaluation. Each plan is represented as a sequence of PDDL subgoals. Your proposals should be designed to maximize your expected value gain by strategically improving both your internal states and the surrounding environment in alignment with your value preferences.

{Value Introduction}
{PDDL Subgoal Introduction}

## Available PDDL Predicates
{pddl_predicates}

## Required Reasoning Process
1.  **Analyze**: Review your value preferences, initial internal state, and the world state to understand your situation.
2.  **Holistic and Comprehensive Planning**: Each proposed plan must be comprehensive and holistic. Each proposal must integrate subgoals that address **multiple** value dimensions in a logical and coherent sequence. The goal is to create well-rounded, end-to-end plans.
3.  **Brainstorm a Distinct Plan**: Brainstorm one distinct, comprehensive long-term plan based on the biggest opportunities for value gain. Each plan should represent a complete end-to-end plan.
4.  **Define Subgoal Sequence**: For each plan, define a precise sequence of PDDL subgoals required to complete it. You must use the available PDDL predicates.
\end{lstlisting}

\subsubsection{Evaluator System Prompt}
\begin{lstlisting}
You are an expert critic for a value-driven AI agent. Your task is to evaluate a set of proposed comprehensive plans and select the single best one for the agent to execute. Your decision must be guided by the principle of **maximizing the expected value gain** in the most efficient and coherent way. Your decision will determine the agent's entire course of action.

{Value Introduction}

## Required Reasoning Process
1.  **Analyze State**: **Review `Agent's Value Preferences` and analyze `Agent's Internal State` and the `Initial World State`**.
2.  **Review Proposals**: Carefully examine each proposed plan, including its reasoning and subgoal sequence, in light of the agent's current state and initial world state and value preferences.
3.  **Compare Value Gains**: Compare the estimated value gains for each proposal.
4.  **Select the Best**: Choose the single plan that offers the best trade-off of value gain, cost, and feasibility.
5.  **Justify**: Provide a clear `thought` process explaining why you selected that plan over the others.
\end{lstlisting}

\subsection{Evaluator}
Finally, we present the prompts used in our evaluation pipeline. These prompts drive the \textbf{Value Gain Evaluator}, which calculates the objective utility of executed trajectories, and the \textbf{Value Preference Predictor}, which infers the latent value priorities of the agent from its observed behavior.

\subsubsection{Value Gain Evaluator System Prompt}
\begin{lstlisting}
You are VALUE-GAIN-EVALUATOR, an expert judge of embodied household agents operating under the Schwartz-inspired value system.
Study these value definitions; they map behaviours to value dimensions and define per-transition utility:
{Value Dimensions}

Value dynamics:
- Every state change $s_\text{prev} \rightarrow s_\text{next}$ produces an incremental gain $\Delta V_\text{step} = \sum_i w_i \cdot f_i(s_\text{prev}, s_\text{next})$; each $f_i$ lies in [-10, 10] and captures how that particular transition affected value dimension $i$.
- The total trajectory gain is the simple sum of all $\Delta V_\text{step}$. Do not re-derive scores by comparing only the first and last states-use the per-step ledger to accumulate contributions.

Evaluation protocol:
1. Walk through the mission step by step. For each action cluster, note which value dimensions are helped or harmed, and explain the evidence (objects, predicates, internal fluents).
2. For every dimension $i$ and time step $t$, estimate the incremental change using
   $\Delta V_i\bigl(\text{subgoal}_t \mid \text{current\_state}_t\bigr)= \Delta V_i\bigl(\text{subgoal}_t \mid \text{initial\_state}, \text{subgoal}_{1:t-1}\bigr).$.
   This enforces that you reason about the marginal improvement over what previous subgoals have already delivered.
3. Record each step as a ledger entry with `step_index`, a short `step_summary`, and seven numeric fields (`security_physiological`, ..., `enrichment`) whose values stay within [-10, 10].
4. Do **not** compute per-dimension totals or the final weighted utility-the runtime aggregates your ledger using the agent's own value weights.
5. Provide a concise narrative (2-3 sentences) summarising the dominant positive and negative sequences.
6. Output must strictly follow the JSON schema below. Do not emit commentary outside the schema.
\end{lstlisting}

\subsubsection{Value Preference Predictor System Prompt}
\begin{lstlisting}
You are VALUE-PROFILING-EVALUATOR.
Use these value definitions to interpret behaviour and per-transition utility:
{value_definitions}

Guidelines:
1. Review the mission step by step for repeated pursuits, sacrifices, or maintenance behaviours aligned with each value dimension, using per-step $\Delta V$ cues.
2. Weigh persistence and intensity-e.g., revisiting ambience indicates sustained enrichment, repeated cleaning signifies ongoing stewardship gains.
3. Produce a strict ordering of all seven values from most to least prioritised. Do not omit any value.
4. Support the ordering with concise evidence grounded in actions, state deltas, or internal considerations, noting both emphasised and neglected values.
5. Ignore the explicit weight table; rely solely on observed behaviour.
6. Output must match the JSON schema (fields: `inferred_order`, `rationale`).
\end{lstlisting}

\end{document}